\documentclass{article}

\usepackage{iclr2027_conference,times}

\usepackage[utf8]{inputenc}
\usepackage[T1]{fontenc}
\usepackage{hyperref}
\usepackage{url}
\usepackage{microtype}
\usepackage{enumitem}
\usepackage{graphicx}
\usepackage{booktabs,array}
\usepackage{tabularx}
\usepackage{multirow}
\usepackage[table]{xcolor}
\usepackage{amsmath}
\usepackage{amssymb}
\usepackage{amsfonts}
\usepackage{mathtools}
\usepackage{amsthm}
\usepackage[capitalize,noabbrev]{cleveref}
\usepackage{wrapfig}
\usepackage{capt-of}
\usepackage{flafter}
\usepackage{float}
\usepackage{placeins}
\newsavebox{\ablationtablebox}
\newsavebox{\componenttablebox}
\newsavebox{\significancetablebox}
\newsavebox{\transfertablebox}
\let\cite\citep
\theoremstyle{plain}
\newtheorem{theorem}{Theorem}[section]
\newtheorem{proposition}[theorem]{Proposition}
\newtheorem{lemma}[theorem]{Lemma}

\theoremstyle{definition}

\theoremstyle{remark}
\newtheorem{remark}[theorem]{Remark}

\title{Graphene: A Lightweight Transformer with Ising Mean-Field Refinement for Image Super-Resolution}

\author{Seungki Hong\qquad Taekwang Jang\\
ETH AI Center\\
ETH Zurich, Switzerland}

\iclrfinalcopy
\definecolor{codeblue}{HTML}{215CAF}
\hypersetup{pdfborder={0 0 0.5},pdftitle={},pdfauthor={},pdfsubject={},pdfkeywords={},pdfcreator={},pdfproducer={}}
\ifdefined\pdfinfoomitdate\pdfinfoomitdate=1\fi
\ifdefined\pdfsuppressptexinfo\pdfsuppressptexinfo=15\fi
\ifdefined\pdftrailerid\pdftrailerid{}\fi

\usepackage{etoolbox}
\makeatletter
\patchcmd{\@maketitle}{\vskip 0.3in minus 0.1in}{\vskip 0.15in minus 0.05in}{}{\errmessage{title spacing patch failed}}
\makeatother

\newcommand{\classicalLatencyReduction}{83.9}
\newcommand{\classicalMemoryReduction}{63.7}
\newcommand{\lightLatencyReduction}{95.9}
\newcommand{\lightMemoryReduction}{92.0}

\begin{document}

\maketitle
\fancyhead[L]{Preprint}

\begin{abstract}
This paper proposes \textit{Graphene}, a lightweight image super-resolution architecture with two complementary feature refinement mechanisms: Ising-model-inspired local feature refinement using an \textit{Ising Refinement Block} (IRB) and stage-history-based feature refinement using a \textit{History Attention} (HistAttn) module. The IRB reuses attention output features and pre-softmax attention logits to construct signed pairwise token couplings within each local window, applies unrolled mean-field updates, and refines features through a gated residual correction. These signed couplings allow alignment and opposition between latent spins to modulate local feature refinement. HistAttn attends over the current stage output and earlier representations at the same token position and adaptively blends the current stage output with a history-weighted feature aggregate without adding spatial pairwise attention. On Urban100 at $\times4$, Graphene and Graphene-light improve over IET and IET-light by \textbf{0.24 dB} and \textbf{0.38 dB}, respectively, with \textbf{34.3\%/13.7\% fewer parameters}, \textbf{44.8\%/44.4\% fewer FLOPs}, \textbf{\classicalLatencyReduction\%/\lightLatencyReduction\% lower latency}, and \textbf{\classicalMemoryReduction\%/\lightMemoryReduction\% lower peak memory}. \textit{Real-world SR} comparisons further show that Graphene recovers fine details and preserves textures across diverse degradations. Code and reference checkpoints are available at \href{https://github.com/skethz/Graphene}{\textcolor[HTML]{215CAF}{github.com/skethz/Graphene}}.
\end{abstract}

\section{Introduction}

\begin{figure}[h!]
\centering
\vspace{-1em}
\includegraphics[width=\textwidth]{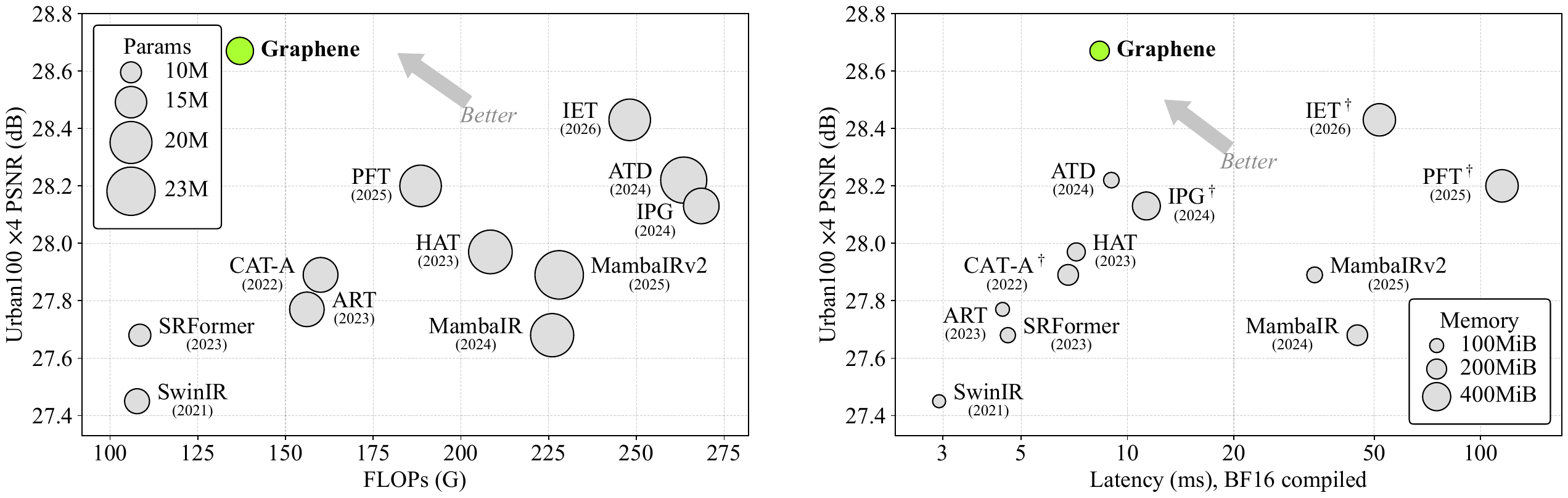}
\vspace{-2.5em}
\caption{Urban100 $\times4$ PSNR versus FLOPs (\textit{left}; circle diameter: parameters) and latency (\textit{right}, log scale; circle area: peak memory). All methods run BF16 compiled inference (BF16 autocast, \texttt{torch.compile} max-autotune, TF32 off) on a GH200 GPU with $64\times64$ LR inputs. $\dagger$: one part of the official code cannot run this setting, so only that part is changed and the best value is reported (CAT-A: \texttt{view} after \texttt{torch.roll}; IPG: first-forward caches; PFT, IET: sparse-attention CUDA operators, in FP32 and FP16). Our released code reproduces all values; FP32 eager results are in Appendix~\ref{app:efficiency}, where Graphene again has the highest PSNR and is faster with less memory than IET.}
\label{fig:bubble}
\end{figure}

Image super-resolution (SR) reconstructs fine image structure from low-resolution (LR) observations. Windowed attention aggregates features within local regions, and recent architectures improve reconstruction by expanding or adapting the features available to each token~\cite{swinir,hat,srformer,mambair,mambairv2,pft,iet}. The challenge is to improve reconstruction while limiting parameter count, FLOPs, latency, and peak memory.

Graphene reuses two quantities already computed in a windowed Transformer: pre-softmax attention logits and intermediate stage outputs. The logits coordinate residual-correction gates within each window; the stored stage outputs supply representations from different depths at the same token position. Separate IRB and HistAttn modules implement these operations within the LR feature-processing trunk.

The \textit{Ising Refinement Block} (IRB) converts logits from Permuted Self-Attention (PSA)~\cite{srformer} into signed, symmetric interactions between latent spins. Three mean-field updates produce magnetizations that gate learned residual corrections to PSA output features. Positive and negative couplings favor aligned and opposing spins, respectively; the resulting gates coordinate correction magnitudes within the processing window.

The \textit{History Attention} (HistAttn) module uses a learned probe to score the current stage output and earlier representations at each token position. Softmax weights form a feature aggregate, which is fused with the current output through a learned scalar. The weights depend on the input features; the aggregation adds no spatial pairwise attention.

Component ablations compare IRB and HistAttn at a fixed window size; coupling ablations vary signs and symmetry; and paired continuation compares intact IRB with a unary-only gate. Transfer experiments integrate IRB into HAT~\cite{hat}. Graphene has the highest reported PSNR in all 15 classical dataset--scale comparisons, and Graphene-light leads 12 of 15 lightweight comparisons, including all five at $\times4$. \Cref{fig:bubble} compares classical SR PSNR against FLOPs and latency, with circle diameters encoding parameter count and circle areas encoding peak memory.

The contributions are:
\begin{itemize}[nosep, leftmargin=*]
  \item This work introduces \textit{IRB}, an Ising-model-inspired refinement block that derives signed local interactions from attention logits and uses finite mean-field updates to gate residual feature corrections, \textit{enabling coordinated refinement of fine local structures}.
  \item This work introduces \textit{HistAttn}, a stage-history attention module that adaptively reuses current and earlier representations at each token position, allowing later stages to retain and \textit{combine useful features from different depths without additional spatial pairwise attention}.
  \item This work achieves state-of-the-art reconstruction quality on \textit{classical, lightweight, and real-world SR benchmarks} while maintaining \textit{low parameter counts, FLOPs, latency, and memory usage}, as demonstrated for classical SR in \Cref{fig:bubble}.
\end{itemize}

\section{Related Work}\label{sec:related_work}

Windowed Transformer SR progresses from SwinIR~\cite{swinir}, which adopts alternating window-based and shifted-window self-attention from the Swin Transformer~\cite{swin}, to HAT~\cite{hat}, which adds channel-attention convolution and overlapping cross-window attention to strengthen local and neighboring-window interactions.
SRFormer~\cite{srformer} introduces Permuted Self-Attention (PSA), which compresses and spatially rearranges keys and values while preserving full-resolution queries to reduce attention cost, and a convolutional feed-forward network (ConvFFN) to recover high-frequency details.
Beyond these windowed Transformers, MambaIR~\cite{mambair} adapts Mamba~\cite{mamba} to SR using four-direction 2D selective scans with local convolution and channel attention; MambaIRv2~\cite{mambairv2} combines state-space modeling, semantic prompts, pixel reordering, local attention, and the SRFormer-style ConvFFN~\cite{srformer} for non-causal restoration; PFT~\cite{pft} progressively filters irrelevant tokens through cross-layer Progressive Focused Attention; and IET~\cite{iet} uses Individualized Exploratory Attention to select content-adaptive candidates for each token.

\section{Methods}\label{sec:methods}

\subsection{Architecture and Design Rationale}

\begin{figure}[!ht]
\centering
\includegraphics[width=\textwidth]{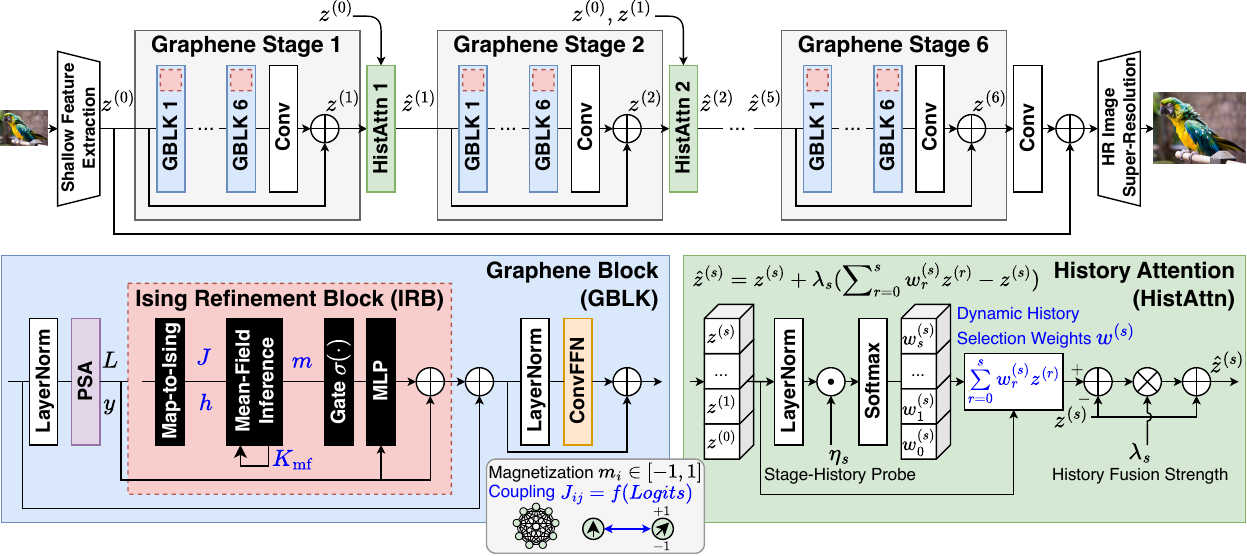}
\vspace{-2em}
\caption{Overall architecture of Graphene. \textit{Top:} Given an LR input, Graphene applies shallow feature extraction with a convolutional stem, deep feature extraction through six Graphene stages, each containing six Graphene Blocks (GBLKs), followed by a trunk convolution, and HR reconstruction with a PixelShuffle head~\cite{pixelshuffle} and a long residual connection. \textit{Bottom-left:} Each GBLK applies LayerNorm, PSA~\cite{srformer}, an Ising Refinement Block (IRB), and a residual LayerNorm--ConvFFN branch. The IRB takes the PSA output features $y$ and pre-softmax logits $L$ as inputs: each token feature $y_i$ is projected to a scalar field $h_i$ and an initial magnetization $m_i^{(0)}$, while the logits $L$ are mapped to a signed, symmetric window-level coupling matrix $J$ with a zero diagonal. Thus, the resulting couplings define a window-local, undirected spin interaction graph. The IRB then performs $K_{\mathrm{mf}}$ mean-field updates to obtain magnetizations $m$, which gate a lightweight residual MLP correction. The final magnetizations modulate token-wise residual corrections through their gates. \textit{Bottom-right:} HistAttn connects successive Graphene stages by matching a learnable stage-history probe $\eta_s$ with LayerNorm-normalized stored features, producing dynamic history selection weights $w_r^{(s)}$ over the stage history $\{z^{(0)},\ldots,z^{(s)}\}$, where $r$ indexes the stored feature. The selected history is then fused with the current stage output $z^{(s)}$ using a learnable history fusion strength $\lambda_s$: $\hat{z}^{(s)}=z^{(s)}+\lambda_s(\sum_{r=0}^{s}w_{r}^{(s)}z^{(r)}-z^{(s)})$. This update refines the current stage output by adaptively blending it with a history-weighted feature aggregate.}
\label{fig:graphene}
\end{figure}

Graphene processes LR features with a convolutional stem, a sequence of residual Graphene stages, a trunk convolution, and a PixelShuffle reconstruction head~\cite{pixelshuffle}. Each Graphene block contains LayerNorm, PSA~\cite{srformer}, an IRB when enabled, and a residual LayerNorm--ConvFFN branch. HistAttn is inserted between successive stages. \Cref{fig:graphene} summarizes the full configuration.

\noindent\textbf{Two refinement axes.} Within each head, PSA aggregates value features with nonnegative attention weights. The IRB reuses the underlying logits to coordinate the strengths of token-wise residual corrections through signed latent-spin interactions, without computing another query--key affinity matrix. Earlier stages provide alternative representations at the same spatial location. HistAttn makes their weighting depend on the token features, while residual fusion preserves a direct path from the current stage output. Together, these operations augment local refinement and depth-wise reuse without global spatial pairwise attention.

\noindent\textbf{Full and lightweight configurations.} Graphene uses six stages, six blocks per stage, 180 channels, and window side length 32. Graphene-light uses four stages, six blocks per stage, 60 channels, and window side length 16. Both use six attention heads, PSA reduction factor $\alpha_{\mathrm{psa}}=2$, and $K_{\mathrm{mf}}=3$ mean-field updates per enabled IRB. IRB is enabled in every full-model block and only in the last block of each lightweight stage. Complete dimensions, initialization, and reconstruction settings are given in Appendix~\ref{app:implementation}.

\subsection{Ising Refinement Block}

\noindent\textbf{Motivation.} PSA's softmax weights are nonnegative and only mix value features, so they cannot express that two tokens should be corrected in opposite ways. IRB reuses the same logits to coordinate how strongly each token is corrected. Centering and symmetrization turn them into signed, reciprocal couplings; positive couplings favor aligned spins and negative couplings opposing ones, so each token's correction strength depends on its partners. The couplings and unary fields define an Ising energy (\Cref{eq:ising}). IRB unrolls its mean-field stationarity condition (\Cref{eq:mean-field}; \Cref{prop:app-stationary}) for $K_{\mathrm{mf}}$ steps, and the final magnetization sets a bounded gate on a token-wise residual correction (\Cref{eq:refine}; \Cref{prop:app-bounds,prop:app-gate}).

\noindent\textbf{Attention logits as local couplings.} Consider a window with $n=W^2$ query tokens and $H$ attention heads. PSA produces logits between query $i$ and reduced key token $u$:
\begin{equation}\label{eq:logits}
L^{(a)}_{iu}=\frac{\langle q^{(a)}_i,k^{(a)}_u\rangle}{\sqrt{d_h}}+b^{(a)}_{iu}+\mathcal M_{iu},
\qquad d_h=C_{\mathrm{feat}}/H,
\end{equation}
where $a$ indexes heads, $C_{\mathrm{feat}}$ is the feature width, $b$ is the relative-position bias, and $\mathcal M$ is the optional shifted-window mask. Queries and keys are projections of LayerNorm-normalized input features. PSA compresses and permutes the key/value branches, leaving $n/\alpha_{\mathrm{psa}}^2$ key/value tokens per window.

Let $\pi(j)$ map fine token $j$ to its reduced key cell. Lifting the logits and averaging heads gives
\begin{equation}\label{eq:lift}
\bar L^{(a)}_{ij}=\frac{L^{(a)}_{i,\pi(j)}}{\alpha_{\mathrm{psa}}^2},\qquad
\tilde J_{ij}=\begin{cases}H^{-1}\sum_{a=1}^{H}\bar L^{(a)}_{ij},&\text{valid pair},\\0,&\text{otherwise}.\end{cases}
\end{equation}
Each reduced key cell is replicated $\alpha_{\mathrm{psa}}^2$ times; the reciprocal factor preserves its summed contribution before subsequent transformations. With $\alpha_{\mathrm{psa}}=2$, each reduced logit gives four copies scaled by $1/4$. Head averaging supplies one interaction graph. Invalid shifted-window pairs are zeroed at this step to remove the attention-mask sentinel values.

The couplings are then centered, symmetrized, and made diagonal-free:
\begin{equation}\label{eq:coupling}
\hat J_{ij}=\tilde J_{ij}-\frac1n\sum_{k=1}^{n}\tilde J_{ik},\qquad
J_{ij}=\frac{\gamma}{2}(\hat J_{ij}+\hat J_{ji})\,\mathbf1[i\ne j],
\end{equation}
where $\gamma>0$ controls coupling strength. Centering expresses affinities relative to each query's mean and permits both signs; symmetrization supplies reciprocal interactions for the undirected Ising energy. The final matrix has no self-couplings, but its row sums need not remain zero. Centering and symmetrization can produce nonzero couplings at previously masked positions. The graph remains confined to the processing window.

\noindent\textbf{Latent spins and finite mean-field inference.} For PSA output feature $y_i\in\mathbb R^{C_{\mathrm{feat}}}$, learned affine projections define a unary field and an initial magnetization:
\begin{equation}\label{eq:fields}
h_i=W_h^\top y_i+b_h,\qquad m_i^{(0)}=\tanh(W_m^\top y_i+b_m).
\end{equation}
The update is motivated by binary spins $\xi_i\in\{-1,+1\}$ with energy
\begin{equation}\label{eq:ising}
E(\boldsymbol\xi)=-\sum_i h_i\xi_i-\frac12\sum_{i,j}J_{ij}\xi_i\xi_j,\qquad
p(\boldsymbol\xi)=Z^{-1}\exp[-\beta E(\boldsymbol\xi)],
\end{equation}
where $\beta>0$ is the inverse temperature and $Z$ is the partition function. Under a factorized distribution $q$, the magnetization $m_i=\mathbb E_q[\xi_i]$ is the mean latent spin. Mean-field stationarity gives the self-consistency equation used by IRB as a differentiable recurrence:
\begin{equation}\label{eq:mean-field}
m_i^{(t+1)}=\tanh\!\left(\beta\left(h_i+\sum_{j=1}^{n}J_{ij}m_j^{(t)}\right)\right),
\qquad t=0,\ldots,K_{\mathrm{mf}}-1.
\end{equation}
Positive and negative couplings give reinforcing and opposing influences between latent states, respectively. The field and coupling matrix remain fixed within each unroll. Three updates per IRB give 108 and 12 updates along the full and lightweight forward paths, respectively, with a separate local system in each block.

\noindent\textbf{Spin-gated residual correction.} The final magnetization controls a token-wise residual MLP:
\begin{equation}\label{eq:refine}
g_i=\sigma(\kappa m_i^{(K_{\mathrm{mf}})}),\qquad
\tilde y_i=y_i+g_iR(y_i),
\end{equation}
where $\sigma$ is the logistic sigmoid, $\kappa$ controls gate scale, and $R$ applies LayerNorm followed by linear--GELU--linear layers. The identity path retains the PSA feature; the residual determines the correction direction, and signed spin interactions modulate its positive gate. For $\kappa=1$, $m_i\in[-1,1]$ gives $g_i\in[0.269,0.731]$ approximately, with $g_i=0.5$ at neutral magnetization. Small learned residuals can preserve homogeneous features, while larger residuals permit stronger refinement. Graphene-light additionally scales $R$ by a learned per-IRB scalar initialized to zero.

The finite unroll is bounded by $\tanh$ and does not require convergence to a fixed point. For a fixed window, $\beta\|J\|_2<1$ is a sufficient contraction condition, but it is not enforced during training. Appendix~\ref{app:formal} gives the mean-field derivation, convergence-rate bound, and proofs verified in Lean~\cite{lean4}.

\subsection{History Attention}

Let $z^{(0)}\in\mathbb R^{B\times N\times C_{\mathrm{feat}}}$ denote shallow features, with batch size $B$ and $N$ spatial tokens. Let $\mathcal G_s$ be stage $s$, and define $\hat z^{(0)}=z^{(0)}$. The stage output is $z^{(s)}=\mathcal G_s(\hat z^{(s-1)})$ for $s=1,\ldots,S$. HistAttn refines this output before the next stage, for $s<S$.

At position $p$, a learned, input-independent probe $\eta_s\in\mathbb R^{C_{\mathrm{feat}}}$ scores the stored features:
\begin{equation}\label{eq:history-score}
e^{(s)}_{r,p}=\frac{\langle\eta_s,\mathrm{LN}(z^{(r)}_p)\rangle}{\sqrt{C_{\mathrm{feat}}}}+\rho\,\mathbf1[r=s],\qquad
w^{(s)}_{r,p}=\frac{\exp(e^{(s)}_{r,p})}{\sum_{\ell=0}^{s}\exp(e^{(s)}_{\ell,p})}.
\end{equation}
The current stage is included in the history, and $\rho$ biases selection toward it. Although the probe is shared across inputs, its scores and selection weights depend on the features at each token position. The aggregate and fused feature are
\begin{equation}\label{eq:history-fusion}
c^{(s)}_p=\sum_{r=0}^{s}w^{(s)}_{r,p}z^{(r)}_p,\qquad
\hat z^{(s)}_p=z^{(s)}_p+\lambda_s(c^{(s)}_p-z^{(s)}_p).
\end{equation}
The weights make $c^{(s)}_p$ a convex history aggregate. The learned scalar $\lambda_s\in\mathbb R$ controls residual fusion: zero preserves the current feature and one selects the aggregate. Because $\lambda_s$ is unconstrained, the complete fusion is affine and may extrapolate. No HistAttn follows the final stage, so the deep feature is $z^{(S)}$.

\noindent\textbf{Computational cost.} For one image, window size $n=W^2$, fixed head count, and fixed MLP ratios, PSA costs $O(NC_{\mathrm{feat}}^2+NnC_{\mathrm{feat}}/\alpha_{\mathrm{psa}}^2)$ per block. IRB adds $O(K_{\mathrm{mf}}Nn+NC_{\mathrm{feat}}^2)$ for the coupling updates and residual MLP. HistAttn processes $s+1$ representations at insertion $s$, giving $O(S^2NC_{\mathrm{feat}})$ total history-fusion cost. Working coupling storage is $O(Nn)$ per IRB and stored history occupies $O(SNC_{\mathrm{feat}})$ feature entries. For fixed windows and architecture, the operation counts and stored coupling and feature entries scale linearly with image area.

\section{Experiments}\label{sec:experiments}

\subsection{Experimental Setup}\label{sec:setup}

\noindent\textbf{Data and evaluation.} The training recipe uses DF2K, comprising DIV2K~\cite{div2k} and Flickr2K~\cite{flickr2k}, with bicubic LR inputs. Classical and lightweight SR are evaluated on Set5~\cite{set5}, Set14~\cite{set14}, BSD100~\cite{bsd100}, Urban100~\cite{urban100}, and Manga109~\cite{manga109}. PSNR/SSIM~\cite{ssim} are computed on the Y channel after removing a border equal to the scale factor. Graphene predictions use EMA weights, with the image conversion and evaluation conventions specified in Appendix~\ref{app:implementation}.

\noindent\textbf{Training.} The base schedule runs for $500{,}000$ iterations on $128\times128$ LR patches with $\ell_1$ loss. Muon~\cite{muon} updates linear-layer matrices in the shared LR encoder; Adam~\cite{adam}, with $(\beta_1,\beta_2)=(0.9,0.999)$, updates the remaining parameters. Both start at $2\times10^{-4}$ and halve at $250{,}000$, $400{,}000$, $450{,}000$, and $475{,}000$ iterations. Training uses random flips and right-angle rotations, BF16 autocast~\cite{mixed} with FP32 parameters and optimizer states, and global batch size 32 across 16 32GB RTX 5090 GPUs. The ablations below use their separately stated budgets.

\noindent\textbf{Efficiency.} Profiling uses $64\times64$ LR inputs at batch size one and counts two FLOPs per multiply--accumulate. Latency and peak allocated CUDA memory are measured for every method with one benchmark script and one software stack: an NVIDIA GH200 GPU with Python 3.12.3, PyTorch 2.9.1, CUDA 13.0, and cuDNN 9.23. All methods use BF16 compiled inference: BF16 autocast with FP32 weights, \texttt{torch.compile} with the Inductor backend in max-autotune mode, and TF32 disabled. Each run starts a fresh process, performs 50 warm-up forward passes, resets the allocator statistics, and times 200 forward passes with CUDA events. The first timed pass re-allocates activation buffers after the reset, so latency is the mean of the remaining 199 passes, and the reported latency is the median over seven runs. Max-autotune replays CUDA graphs, whose memory pool is not visible to the allocator statistics, so peak memory is the median of three runs of the same compilation without CUDA graphs (max-autotune-no-cudagraphs).

For four methods, one part of the official code cannot run this setting, and only that part is changed. In CAT-A, Inductor's lowering of \texttt{torch.roll} returns non-contiguous strides, so four \texttt{view} calls in each shifted block fail to compile; they become \texttt{reshape}, which returns the same values. IPG caches its graph-construction constants and position-bias table at the first forward pass; under CUDA graphs, the next replay overwrites these caches, so one eager pass with the same autocast setting fills them before the first compiled call. The sparse-attention CUDA operators of PFT and IET accept FP32 (IET also FP16) but not BF16; these calls run in FP32 for PFT and in FP16 for IET, and all other operations remain in BF16 autocast. Where alternatives exist, they were also measured: CAT-A's shifted blocks run eagerly, IPG without CUDA graphs, IET's sparse operators in FP32, and NATTEN's GEMM kernels with TF32 disabled. The fastest variant is reported.

The PSNR axis of \Cref{fig:bubble} shows the values of \Cref{tab:sr_benchmark}; BF16 compiled inference changes Graphene's Urban100 $\times4$ PSNR by less than 0.001 dB. \Cref{tab:large_input} extends the comparison to larger LR inputs, and Appendix~\ref{app:efficiency} reports FP32 eager latency and peak memory for every method.

\subsection{Ablation of IRB}

The ablations use DIV2K, $10{,}000$ training iterations, and Urban100 $\times2$ evaluation, with the full backbone and three mean-field updates. The controls in \Cref{tab:component_ablation,tab:history_selection} train from scratch under this protocol, matching shared-parameter initialization, minibatches, and optimization settings.

{\setlength{\intextsep}{4pt}%
\begin{table}[H]
\caption{Component baselines and IRB interactions on Urban100 at $\times2$. Panel (b) retains HistAttn and the residual MLP at window side length 32; $\Delta$ is relative to full Graphene.}
\label{tab:component_ablation}
\centering
\begin{tabular*}{\columnwidth}{@{\extracolsep{\fill}}cccrrrr@{}}
\toprule
\multicolumn{7}{l}{\textit{(a) Component baselines}} \\
Window & IRB & HistAttn & Params (M) & FLOPs (G) & PSNR (dB) & $\Delta$ (dB) \\
\midrule
16 & -- & -- & 10.3 & 91.0 & 30.14 & $-0.79$ \\
32 & -- & -- & 10.5 & 111.4 & 30.33 & $-0.60$ \\
32 & -- & Enabled & 10.5 & 111.6 & 30.53 & $-0.40$ \\
32 & Enabled & -- & 12.9 & 131.8 & 30.88 & $-0.05$ \\
32 & Enabled & Enabled & 12.9 & 132.0 & \textbf{30.93} & $0.00$ \\
\bottomrule
\addlinespace
\toprule
\multicolumn{7}{l}{\textit{(b) Gate and coupling controls}} \\
\multicolumn{3}{l}{IRB configuration} & Params (M) & FLOPs (G) & PSNR (dB) & $\Delta$ (dB) \\
\midrule
\multicolumn{3}{l}{Constant gate ($g_i=0.5$)} & 12.8 & 130.9 & 30.85 & $-0.08$ \\
\multicolumn{3}{l}{Unary-only gate ($J=0$)} & 12.9 & 131.0 & 30.89 & $-0.04$ \\
\multicolumn{3}{l}{Nonnegative, symmetric} & 12.9 & 132.0 & 30.90 & $-0.03$ \\
\multicolumn{3}{l}{Signed, asymmetric} & 12.9 & 132.0 & 30.86 & $-0.07$ \\
\multicolumn{3}{l}{Signed, symmetric (full IRB)} & 12.9 & 132.0 & \textbf{30.93} & $0.00$ \\
\bottomrule
\end{tabular*}
\end{table}}

\noindent\textbf{Module contributions.} At window side length 32 (\Cref{tab:component_ablation}a), IRB improves PSNR from $30.33$ to $30.88$ dB, while HistAttn alone reaches $30.53$ dB. Their combination reaches $30.93$ dB, adding $0.60$ dB to the same-window baseline. IRB adds $2.4$M parameters and $20.4$G FLOPs; HistAttn adds approximately $0.2$G FLOPs. Increasing the baseline window from 16 to 32 separately adds $0.19$ dB.

\noindent\textbf{Interactions beyond residual capacity.} The controls in \Cref{tab:component_ablation}b retain the residual MLP and HistAttn. The constant gate fixes $g_i=0.5$; the unary-only gate sets $J=0$ while retaining the learned field. Unary gating improves PSNR from $30.85$ to $30.89$ dB, and full IRB adds $0.04$ dB to reach $30.93$ dB. These differences support feature-dependent gating and pairwise coordination beyond the added residual MLP.

\noindent\textbf{Signed, symmetric couplings.} The controls in \Cref{tab:component_ablation}b give $30.90$ dB with nonnegative couplings and $30.86$ dB without symmetry, respectively $0.03$ and $0.07$ dB below full IRB. These results favor signed, symmetric interactions. Three mean-field updates also give the highest PSNR among the tested unroll lengths (Section~\ref{app:unroll}).

\noindent\textbf{Complementary evidence.} Appendix~\ref{app:additional-experiments} reports six-seed repeatability of the full model ($30.92\pm0.08$ dB, mean $\pm$ sample SD; Section~\ref{app:seed-variation}), transfer of IRB to HAT (Section~\ref{app:irb-transferability}), and paired continuation and regional diagnostics (Section~\ref{app:paired}).

\subsection{Ablation of HistAttn}

The controls in \Cref{tab:history_selection} retain full IRB. A learned shallow skip sets $c_p^{(s)}=z_p^{(0)}$ in \Cref{eq:history-fusion}. Static selection replaces the feature-dependent scores with learned stage-specific logits shared across images and tokens. It accesses the same history as HistAttn, including the current stage. All enabled variants learn $\lambda_s$ at the same stage insertions; Section~\ref{app:unroll} gives the control equations.

\begin{table}[!hb]
\caption{Adaptive history selection on Urban100 at $\times2$ with full IRB and window side length 32. $\Delta$ is relative to HistAttn.}
\label{tab:history_selection}
\centering
\begin{tabular*}{\columnwidth}{@{\extracolsep{\fill}}lrrrr@{}}
\toprule
History configuration & Params (M) & FLOPs (G) & PSNR (dB) & $\Delta$ (dB) \\
\midrule
No history fusion & 12.9 & 131.8 & 30.88 & $-0.05$ \\
Learned shallow skip & 12.9 & 131.8 & 30.89 & $-0.04$ \\
Static history weights & 12.9 & 131.8 & 30.90 & $-0.03$ \\
HistAttn & 12.9 & 132.0 & \textbf{30.93} & $0.00$ \\
\bottomrule
\end{tabular*}
\end{table}

\noindent\textbf{Selection beyond a skip connection.} HistAttn improves PSNR from $30.88$ to $30.93$ dB (\Cref{tab:history_selection}). A learned shallow skip reaches $30.89$ dB, while static history weights give $30.90$ dB. HistAttn's additional $0.04$ and $0.03$ dB support selecting representations according to token features beyond a shallow skip or fixed depth mixture, without spatial pairwise attention.

\subsection{Activation and History Analysis}

\noindent\textbf{Activation patterns.} In \Cref{fig:feature}, the IRB row shows localized contrasts around facade and window-frame structures, while the HistAttn row changes the distribution of responses across the facade with depth. These are observations of relative spatial patterns: each panel is normalized independently, and the Graphene rows disable one module at inference. They complement the retrained component comparisons in \Cref{tab:component_ablation}.

{\setlength{\intextsep}{4pt}%
\begin{figure}[H]
\centering
\includegraphics[width=\textwidth]{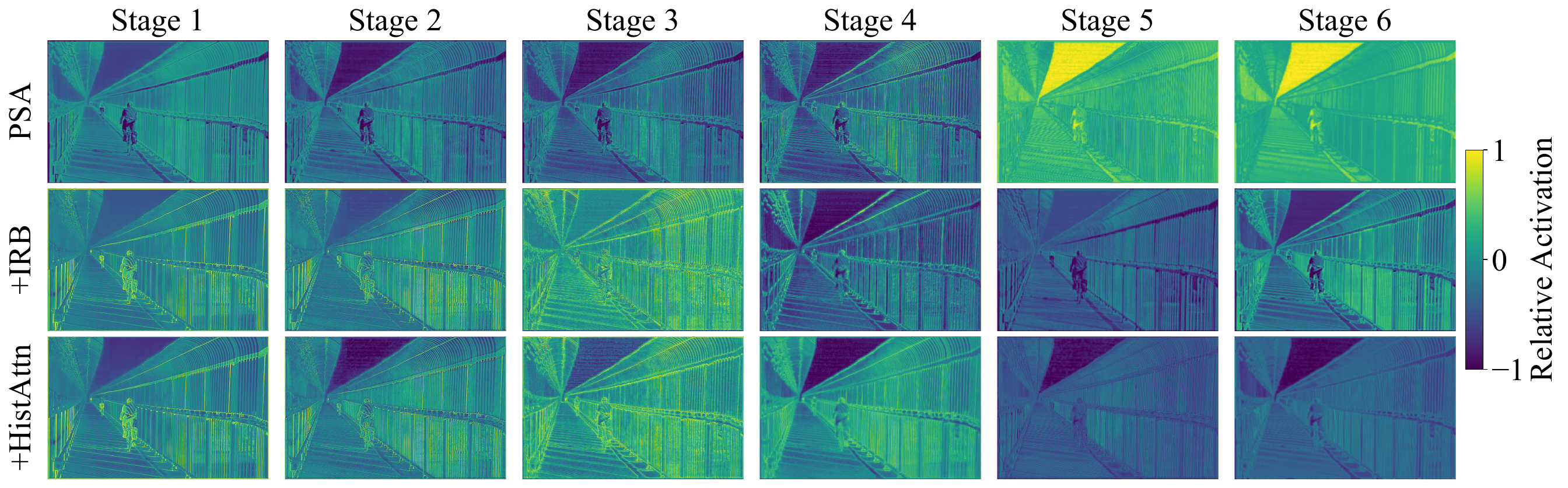}
\vspace{-2.5em}
\caption{Stage-wise channel-mean activation maps for Urban100 img\_024 at $\times2$. The PSA row uses SRFormer; the other rows retain IRB or HistAttn while disabling the other module in Graphene at inference. Each panel is normalized independently, showing relative spatial organization across stages.}
\label{fig:feature}
\end{figure}}

\noindent\textbf{Structure-dependent interactions and history selection.} \Cref{fig:irb_history_analysis} summarizes the measured couplings and history selection. Across $18{,}221$ LR windows, the mean absolute coupling in the selected edge and texture categories is $6.3\times$ and $7.8\times$ that in the flat category, using unrounded category means. HistAttn modules feeding Stages~3--4 assign $96.9$--$98.8\%$ of their selection weight to the shallow feature. The insertion feeding Stage~6 assigns $40.8\%$ to the current stage output and $59.2\%$ to earlier features, before residual fusion. Appendix~\ref{app:features} provides the category definitions, measurement protocols, and cross-scale comparisons.

{\setlength{\intextsep}{4pt}%
\begin{figure}[H]
\centering
\includegraphics[width=\textwidth]{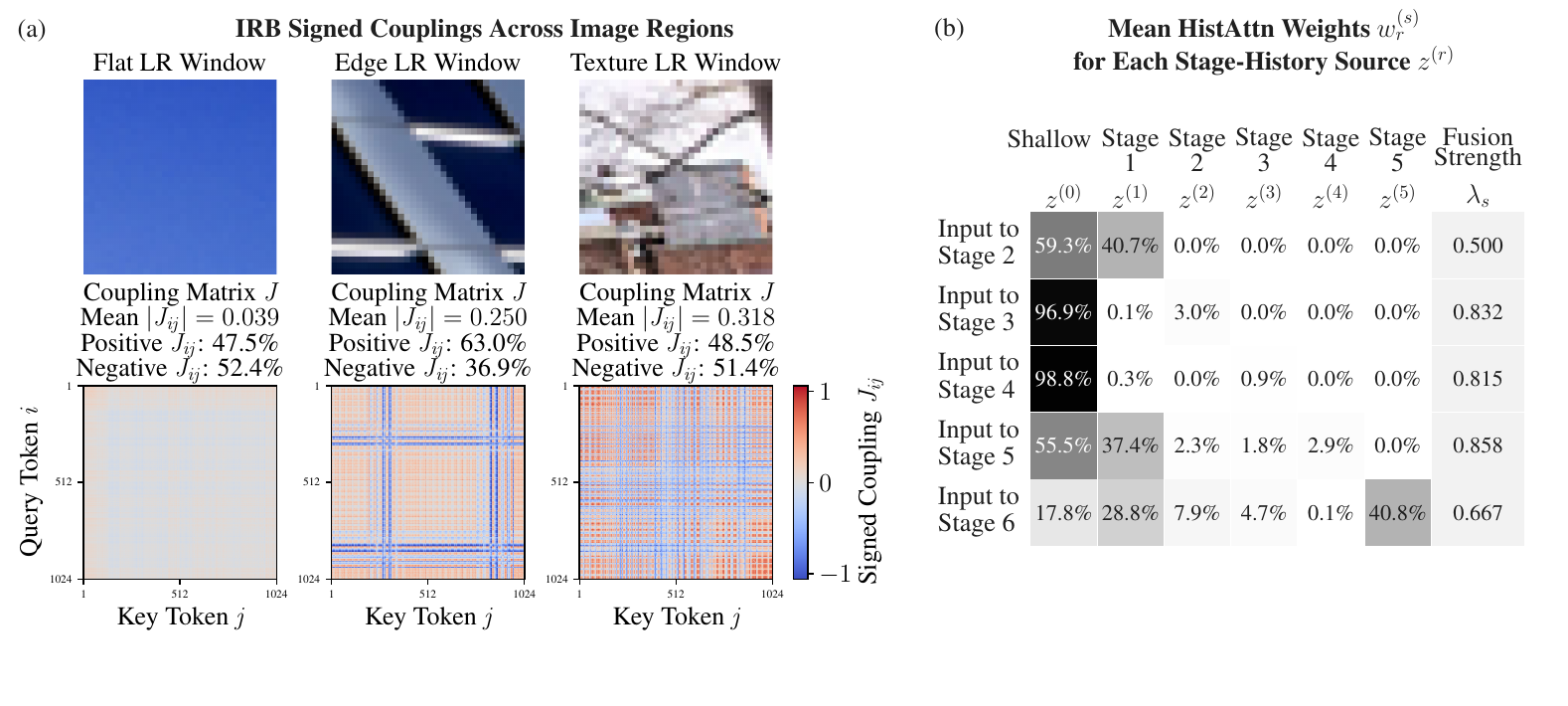}
\vspace{-4em}
\caption{IRB and HistAttn analysis on Urban100 at $\times2$. \textit{Left:} Representative flat, edge, and texture windows and their signed coupling matrices. \textit{Right:} Mean history selection weights before residual fusion, together with the learned fusion strengths. Individual-window values are distinct from category averages.}
\label{fig:irb_history_analysis}
\end{figure}}

\subsection{Parameters, FLOPs, Latency, Peak Memory, and PSNR}

\providecommand{\tbd}{\textcolor{red}{TBD}}
\begin{table}[!t]
\vspace{-0.7em}
\caption{Classical and lightweight $\times4$ SR. Best and second-best PSNR values within each group are bold and underlined, with ties treated equally. Graphene's parameter counts, FLOPs, latency, and peak memory are also bolded. Latency and memory use BF16 compiled inference for all methods ($64\times64$ LR inputs). Full PSNR/SSIM comparisons at $\times2$, $\times3$, and $\times4$ are in Appendix~\ref{app:benchmarks}.}
\label{tab:sr_benchmark}
\centering
\small
\setlength{\tabcolsep}{2.2pt}
\begin{tabularx}{\columnwidth}{@{}l
  *{4}{>{\hsize=1.2\hsize\linewidth=\hsize\raggedleft\arraybackslash}X}
  *{5}{>{\hsize=0.84\hsize\linewidth=\hsize\raggedleft\arraybackslash}X}@{}}
\toprule
\multicolumn{10}{c}{\textbf{Classical SR}} \\
\midrule
Method & Params & FLOPs & Latency & Memory & Set5 & Set14 & B100 & U100 & M109 \\
& (M) & (G) & (ms) & (MiB) & & & & & \\
\midrule
SwinIR~\cite{swinir} & 11.9 & 107.7 & 2.93 & 86.3 & 32.92 & 29.09 & 27.92 & 27.45 & 32.03 \\
CAT-A~\cite{cat} & 16.6 & 160.0 & 6.79 & 216.4 & 33.08 & 29.18 & 27.99 & 27.89 & 32.39 \\
ART~\cite{art} & 16.5 & 156.1 & 4.43 & 97.3 & 33.04 & 29.16 & 27.97 & 27.77 & 32.31 \\
HAT~\cite{hat} & 20.8 & 208.4 & 7.16 & 163.3 & 33.04 & 29.23 & 28.00 & 27.97 & 32.48 \\
SRFormer~\cite{srformer} & 10.5 & 108.5 & 4.58 & 118.2 & 32.93 & 29.08 & 27.94 & 27.68 & 32.21 \\
IPG~\cite{ipg} & 17.0 & 268.5 & 11.31 & 394.8 & 33.15 & 29.24 & 27.99 & 28.13 & 32.53 \\
ATD~\cite{atd} & 22.0 & 263.5 & 9.00 & 123.0 & 33.14 & 29.25 & 28.02 & 28.22 & 32.65 \\
MambaIR~\cite{mambair} & 20.6 & 226.0 & 44.76 & 214.1 & 33.03 & 29.20 & 27.98 & 27.68 & 32.32 \\
MambaIRv2~\cite{mambairv2} & 23.1 & 228.0 & 33.88 & 126.7 & 33.14 & 29.23 & 28.00 & 27.89 & 32.57 \\
PFT~\cite{pft} & 19.8 & 188.5 & 114.99 & 533.2 & 33.15 & 29.29 & 28.02 & 28.20 & 32.63 \\
IET~\cite{iet} & 19.8 & 248.1 & 51.68 & 530.4 & \underline{33.22} & \underline{29.35} & \underline{28.06} & \underline{28.43} & \underline{32.81} \\
\rowcolor{green!5}
\textbf{Graphene} & \textbf{13.0} & \textbf{137.0} & \textbf{8.34} & \textbf{192.6} & \textbf{33.27} & \textbf{29.43} & \textbf{28.12} & \textbf{28.67} & \textbf{32.89} \\
\midrule
\multicolumn{10}{c}{\textbf{Lightweight SR}} \\
\midrule
Method & Params & FLOPs & Latency & Memory & Set5 & Set14 & B100 & U100 & M109 \\
& (K) & (G) & (ms) & (MiB) & & & & & \\
\midrule
SwinIR~\cite{swinir} & 930.0 & 8.8 & 1.12 & 26.8 & 32.44 & 28.77 & 27.69 & 26.47 & 30.92 \\
SRFormer~\cite{srformer} & 873.0 & 8.4 & 1.53 & 28.8 & 32.51 & 28.82 & 27.73 & 26.67 & 31.17 \\
ATD~\cite{atd} & 777.0 & 13.7 & 3.80 & 19.8 & 32.63 & 28.89 & 27.79 & 26.97 & 31.48 \\
MambaIR~\cite{mambair} & 924.0 & 12.0 & 28.32 & 25.5 & 32.51 & 28.85 & 27.75 & 26.75 & 31.26 \\
MambaIRv2~\cite{mambairv2} & 790.0 & 10.1 & 18.46 & 19.9 & 32.51 & 28.84 & 27.75 & 26.82 & 31.24 \\
PFT~\cite{pft} & 792.0 & 10.3 & 45.12 & 294.8 & 32.63 & 28.92 & 27.79 & 27.20 & 31.51 \\
UCAN~\cite{ucan} & 705.0 & 11.3 & 20.47 & 146.0 & 32.65 & 28.95 & 27.79 & 26.89 & 31.50 \\
IET~\cite{iet} & 801.0 & 12.6 & 38.52 & 258.8 & \underline{32.70} & \underline{29.00} & \underline{27.85} & \underline{27.41} & \underline{31.73} \\
\rowcolor{green!5}
\textbf{Graphene-light} & \textbf{691.6} & \textbf{7.0} & \textbf{1.60} & \textbf{20.8} & \textbf{32.80} & \textbf{29.19} & \textbf{27.94} & \textbf{27.79} & \textbf{31.93} \\
\bottomrule
\end{tabularx}
\end{table}

\begin{table}[!t]
\caption{$\times4$ BF16 compiled latency and peak memory at larger LR inputs (width$\times$height), medians over three runs of 5 warm-up and 20 timed passes; OOM: out of GPU memory.}
\label{tab:large_input}
\centering
\small
\begin{tabular}{lrrrr}
\toprule
& \multicolumn{2}{c}{$320\times180$} & \multicolumn{2}{c}{$1280\times720$} \\
\cmidrule(lr){2-3}\cmidrule(lr){4-5}
Method & Latency (ms) & Memory (GiB) & Latency (s) & Memory (GiB) \\
\midrule
\multicolumn{5}{c}{\textbf{Classical SR}} \\
\midrule
SwinIR~\cite{swinir} & 19.6 & 0.37 & 0.254 & 4.65 \\
CAT-A~\cite{cat} & 86.6 & 2.22 & 5.268 & 109.51 \\
ART~\cite{art} & 225.2 & 3.17 & OOM & OOM \\
HAT~\cite{hat} & 49.3 & 0.84 & 0.676 & 11.32 \\
SRFormer~\cite{srformer} & 36.8 & 0.47 & 0.541 & 5.71 \\
IPG~\cite{ipg} & 131.8 & 3.40 & 1.978 & 50.26 \\
ATD~\cite{atd} & 58.8 & 0.62 & 0.733 & 8.18 \\
MambaIR~\cite{mambair} & 192.7 & 1.47 & 2.935 & 20.28 \\
MambaIRv2~\cite{mambairv2} & 116.7 & 0.63 & 1.534 & 8.24 \\
PFT~\cite{pft} & 239.8 & 5.66 & 4.035 & 84.64 \\
IET~\cite{iet} & 484.9 & 3.53 & 8.344 & 51.90 \\
\rowcolor{green!5}
\textbf{Graphene} & \textbf{76.9} & \textbf{1.08} & \textbf{1.145} & \textbf{14.78} \\
\midrule
\multicolumn{5}{c}{\textbf{Lightweight SR}} \\
\midrule
SwinIR~\cite{swinir} & 5.6 & 0.17 & 0.078 & 2.64 \\
SRFormer~\cite{srformer} & 6.8 & 0.18 & 0.089 & 2.45 \\
ATD~\cite{atd} & 15.2 & 0.24 & 0.165 & 3.57 \\
MambaIR~\cite{mambair} & 28.8 & 0.32 & 0.419 & 4.76 \\
MambaIRv2~\cite{mambairv2} & 22.5 & 0.24 & 0.292 & 3.59 \\
PFT~\cite{pft} & 73.5 & 3.66 & 1.201 & 55.36 \\
UCAN~\cite{ucan} & 28.3 & 1.64 & 0.255 & 23.74 \\
IET~\cite{iet} & 147.1 & 1.81 & 2.203 & 26.82 \\
\rowcolor{green!5}
\textbf{Graphene-light} & \textbf{6.8} & \textbf{0.24} & \textbf{0.091} & \textbf{3.55} \\
\bottomrule
\end{tabular}
\end{table}

\noindent\textbf{Classical SR.} \Cref{tab:sr_benchmark} compares $\times4$ PSNR, parameter counts, FLOPs, latency, and peak memory. Graphene has the highest reported PSNR on all five datasets, extending to all 15 dataset--scale settings in Appendix~\ref{app:benchmarks}. On Urban100 at $\times4$, Graphene exceeds IET~\cite{iet} by \textbf{0.24 dB}, using \textbf{34.3\% fewer parameters}, \textbf{44.8\% fewer FLOPs}, \textbf{\classicalLatencyReduction\% lower latency}, and \textbf{\classicalMemoryReduction\% lower peak memory}. \Cref{fig:bubble} visualizes these comparisons. Baselines with lower latency or peak memory than Graphene have at least $0.70$ or $0.45$ dB lower Urban100 PSNR, respectively.

\begin{figure}[!b]
\centering
\includegraphics[width=\textwidth]{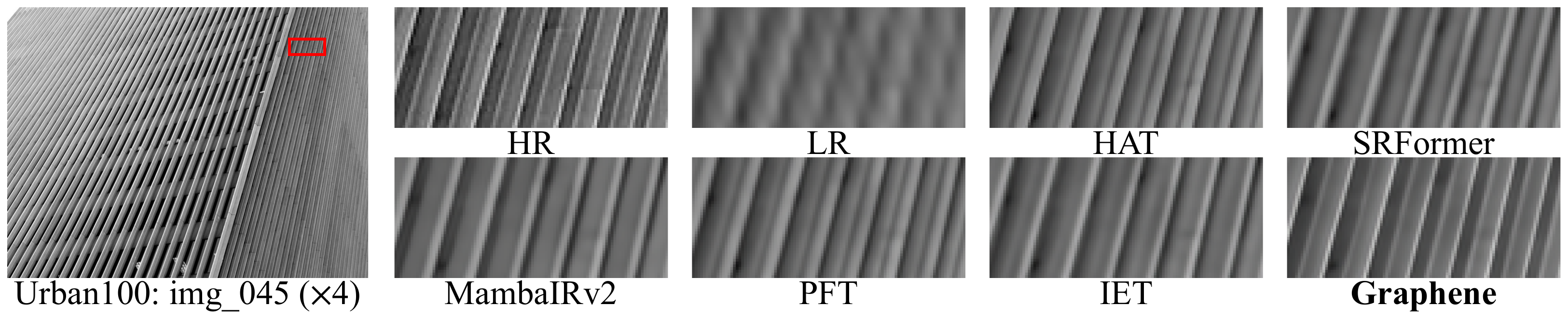}
\includegraphics[width=\textwidth]{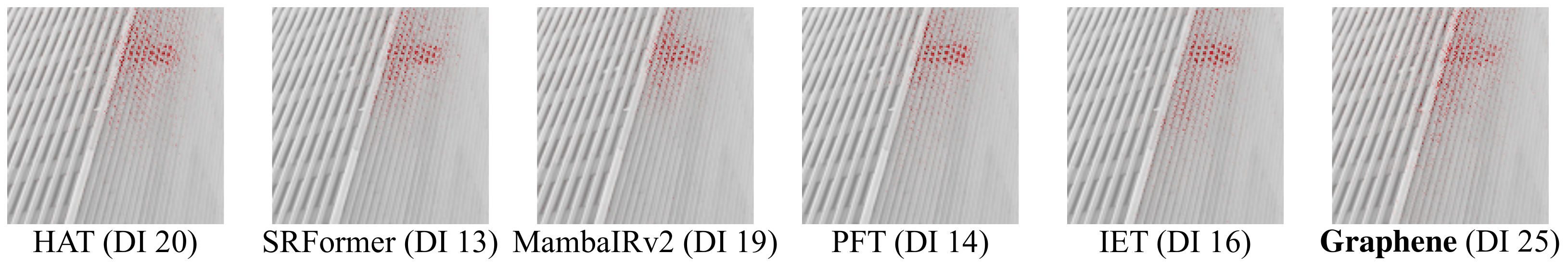}
\vspace{-2em}
\caption{\textit{Top:} $\times4$ SR on Urban100 img\_045. Graphene preserves the repeated line structure visible in the HR crop. \textit{Bottom:} Local Attribution Maps for the outlined output region, with Diffusion Index (DI) values rounded to integers.}
\label{fig:u100_lam}
\end{figure}

\begin{figure}[!t]
\centering
\includegraphics[width=\textwidth]{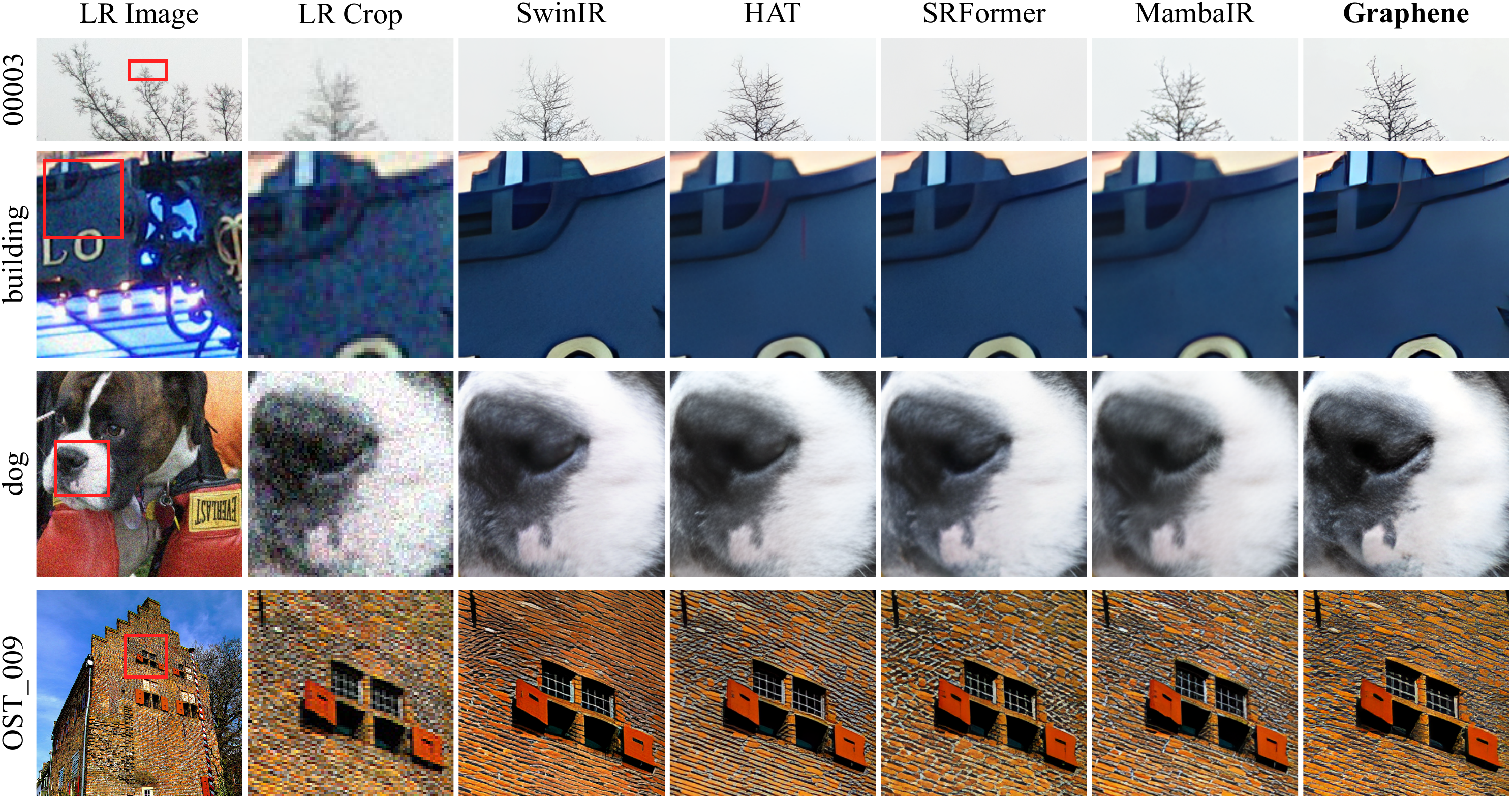}
\vspace{-2em}
\caption{Qualitative $\times4$ \textbf{real-world SR} on representative RealSRSet+5images examples. Graphene reconstructs fine details and textures in the displayed regions.}
\label{fig:realsr}
\end{figure}

\noindent\textbf{Lightweight SR.} The lower panel compares lightweight models, including UCAN~\cite{ucan}. Graphene-light achieves the highest PSNR on the five $\times4$ datasets and on 12 of the 15 settings across scales. Its Urban100 $\times4$ gain over IET-light is \textbf{0.38 dB}, with \textbf{13.7\% fewer parameters}, \textbf{44.4\% fewer FLOPs}, \textbf{\lightLatencyReduction\% lower latency}, and \textbf{\lightMemoryReduction\% lower peak memory}. SwinIR-light and SRFormer-light are faster, and ATD-light and MambaIRv2-light use less memory, with at least $0.82$ dB lower Urban100 PSNR. The four-stage Graphene-light model achieves these $\times4$ PSNR values with $691.6$K parameters.

\noindent\textbf{Visual reconstruction and attribution.} \Cref{fig:u100_lam} compares repeated structures in Urban100 img\_045. Local Attribution Maps (LAM)~\cite{lam} characterize the input locations associated with reconstruction of the selected output patch. Graphene obtains the highest displayed Diffusion Index (DI), indicating the broadest spatial attribution for this region. This measurement characterizes the selected prediction; it does not establish a dataset-wide explanation of the PSNR advantage. Section~\ref{app:attribution} specifies the attribution objective and protocol.

\noindent\textbf{Real-world SR.} \Cref{fig:realsr} compares Graphene with SwinIR, HAT, SRFormer, and MambaIR, the compared baselines providing real-world SR models. Graphene follows SwinIR's real-world training pipeline~\cite{swinir}. The naturally degraded images have no aligned HR references, so the comparison is qualitative, as in SwinIR~\cite{swinir} and MambaIR~\cite{mambair}. The displayed examples illustrate detail and texture recovery across different image content.

\noindent\textbf{Larger LR inputs.} In \Cref{tab:large_input}, Graphene and Graphene-light are $6.3$--$24.3\times$ faster than IET and IET-light with $3.3$--$7.6\times$ less memory; faster or leaner models are $\geq0.45$ dB lower on Urban100, also in FP32 eager inference (Appendix~\ref{app:efficiency}, which also gives the code variants).

\noindent\textbf{Limitations.} The IRB requires materialized pre-softmax attention logits, limiting compatibility with FlashAttention-style kernels~\cite{flashattention}. Nevertheless, Graphene achieves lower inference latency than several state-of-the-art SR methods, demonstrating practical efficiency.

\section{Conclusion}

This paper introduces \textit{Graphene}, a lightweight SR architecture combining Ising-model-inspired local refinement with stage-history reuse. The IRB uses signed, window-local couplings and mean-field updates to gate residual corrections, while HistAttn reuses earlier features at each token position without additional spatial pairwise attention. On Urban100 at $\times4$, Graphene and Graphene-light achieve higher PSNR than the corresponding IET variants with fewer parameters and FLOPs, lower latency, and lower peak memory.

\newpage
\subsection*{AI Use Statement}
Generative AI tools were used to assist with writing and checking formatting
compliance. The authors reviewed all AI-assisted changes and take responsibility
for the final content of this work, including text, claims, and artifacts
produced with the aid of generative AI.

\subsection*{Ethics Statement}
This work uses public image super-resolution datasets and standard benchmark
protocols and does not involve human subjects or private data. The authors have
read and commit to the ICLR Code of Ethics.

\subsection*{Reproducibility Statement}
\Cref{sec:methods} specifies the refinement equations and full/light architecture sizes, and \Cref{sec:experiments} states the benchmark and ablation protocols. Appendix~\ref{app:formal} provides mathematical derivations and the scope of the companion Lean development; Appendix~\ref{app:implementation} gives architecture and evaluation details; Appendix~\ref{app:additional-experiments} provides extended ablations and the paired-continuation protocol; Appendix~\ref{app:features} describes feature and attribution measurements; and Appendix~\ref{app:benchmarks} contains the full comparison tables.
Code and reference checkpoints are available at \href{https://github.com/skethz/Graphene}{\textcolor[HTML]{215CAF}{github.com/skethz/Graphene}}.

\bibliography{formal_refs,ref}

@misc{graphenelean4,
  author = {{Lean project}},
  title = {Lean 4, Version 4.33.1},
  year = {2026},
  howpublished = {\href{https://github.com/leanprover/lean4/releases/tag/v4.33.1}{Official versioned release}}
}

@misc{graphenemathlib4,
  author = {{The mathlib community}},
  title = {mathlib4, Version 4.33.1},
  year = {2026},
  howpublished = {\href{https://github.com/leanprover-community/mathlib4/tree/v4.33.1}{Official versioned source}}
}

@INPROCEEDINGS{div2k,
  author={Timofte, Radu and Agustsson, Eirikur and Gool, Luc Van and Yang, Ming-Hsuan and Zhang, Lei and Lim, Bee and Son, Sanghyun and Kim, Heewon and Nah, Seungjun and Lee, Kyoung Mu and Wang, Xintao and Tian, Yapeng and Yu, Ke and Zhang, Yulun and Wu, Shixiang and Dong, Chao and Lin, Liang and Qiao, Yu and Loy, Chen Change and Bae, Woong and Yoo, Jaejun and Han, Yoseob and Ye, Jong Chul and Choi, Jae-Seok and Kim, Munchurl and Fan, Yuchen and Yu, Jiahui and Han, Wei and Liu, Ding and Yu, Haichao and Wang, Zhangyang and Shi, Honghui and Wang, Xinchao and Huang, Thomas S. and Chen, Yunjin and Zhang, Kai and Zuo, Wangmeng and Tang, Zhimin and Luo, Linkai and Li, Shaohui and Fu, Min and Cao, Lei and Heng, Wen and Bui, Giang and Le, Truc and Duan, Ye and Tao, Dacheng and Wang, Ruxin and Lin, Xu and Pang, Jianxin and Xu, Jinchang and Zhao, Yu and Xu, Xiangyu and Pan, Jinshan and Sun, Deqing and Zhang, Yujin and Song, Xibin and Dai, Yuchao and Qin, Xueying and Huynh, Xuan-Phung and Guo, Tiantong and Mousavi, Hojjat Seyed and Vu, Tiep Huu and Monga, Vishal and Cruz, Cristovao and Egiazarian, Karen and Katkovnik, Vladimir and Mehta, Rakesh and Jain, Arnav Kumar and Agarwalla, Abhinav and Praveen, Ch V. Sai and Zhou, Ruofan and Wen, Hongdiao and Zhu, Che and Xia, Zhiqiang and Wang, Zhengtao and Guo, Qi},
  booktitle={2017 IEEE Conference on Computer Vision and Pattern Recognition Workshops (CVPRW)}, 
  title={{NTIRE 2017 Challenge on Single Image Super-Resolution: Methods and Results}}, 
  year={2017},
  volume={},
  number={},
  pages={1110-1121},
  doi={10.1109/CVPRW.2017.149}}

@INPROCEEDINGS{flickr2k,
  author={Lim, Bee and Son, Sanghyun and Kim, Heewon and Nah, Seungjun and Lee, Kyoung Mu},
  booktitle={2017 IEEE Conference on Computer Vision and Pattern Recognition Workshops (CVPRW)}, 
  title={{Enhanced Deep Residual Networks for Single Image Super-Resolution}}, 
  year={2017},
  volume={},
  number={},
  pages={1132-1140},
  doi={10.1109/CVPRW.2017.151}}

@inproceedings{set5,
	title={{Low-Complexity Single-Image Super-Resolution based on Nonnegative Neighbor Embedding}},
	author={Marco Bevilacqua and Aline Roumy and Christine Guillemot and Marie-line Alberi Morel},
	year={2012},
	pages={135.1--135.10},
	booktitle={Proceedings of the British Machine Vision Conference},
	publisher={BMVA Press},
	editors={Bowden, Richard and Collomosse, John and Mikolajczyk, Krystian},
	isbn={1-901725-46-4},
	doi={10.5244/C.26.135}
}

@inproceedings{set14,
  title={{On Single Image Scale-Up Using Sparse-Representations}},
  author={Zeyde, Roman and Elad, Michael and Protter, Matan},
  booktitle={International Conference on Curves and Surfaces},
  pages={711--730},
  year={2010},
  organization={Springer}
}

@inproceedings{bsd100,
  author={Martin, D. and Fowlkes, C. and Tal, D. and Malik, J.},
  booktitle={Proceedings Eighth IEEE International Conference on Computer Vision (ICCV)}, 
  title={{A Database of Human Segmented Natural Images and its Application to
Evaluating Segmentation Algorithms and Measuring Ecological Statistics}}, 
  year={2001},
  volume={2},
  number={},
  pages={416-423 vol.2},
  doi={10.1109/ICCV.2001.937655}
}

@inproceedings{urban100,
  author={Huang, Jia-Bin and Singh, Abhishek and Ahuja, Narendra},
  booktitle={2015 IEEE Conference on Computer Vision and Pattern Recognition (CVPR)}, 
  title={{Single Image Super-resolution from Transformed Self-Exemplars}}, 
  year={2015},
  volume={},
  number={},
  pages={5197-5206},
  doi={10.1109/CVPR.2015.7299156}
}

@article{manga109,
  title={{Sketch-based Manga Retrieval using Manga109 Dataset}},
  author={Matsui, Yusuke and Ito, Kota and Aramaki, Yuji and Fujimoto, Azuma and Ogawa, Toru and Yamasaki, Toshihiko and Aizawa, Kiyoharu},
  journal={Multimedia Tools and Applications},
  volume={76},
  number={20},
  pages={21811--21838},
  year={2017},
  publisher={Springer}
}

@ARTICLE{ssim,
  author={Zhou Wang and Bovik, A.C. and Sheikh, H.R. and Simoncelli, E.P.},
  journal={IEEE Transactions on Image Processing (TIP)}, 
  title={{Image Quality Assessment: From Error Visibility to Structural Similarity}}, 
  year={2004},
  volume={13},
  number={4},
  pages={600-612},
  doi={10.1109/TIP.2003.819861}
}

@inproceedings{lean4,
  author = {de Moura, Leonardo and Ullrich, Sebastian},
  title = {{The Lean 4 Theorem Prover and Programming Language}},
  booktitle = {Automated Deduction -- CADE 28},
  series = {Lecture Notes in Computer Science},
  volume = {12699},
  pages = {625--635},
  year = {2021},
  publisher = {Springer},
  doi = {10.1007/978-3-030-79876-5_37}
}

@INPROCEEDINGS{swin,
  author={Liu, Ze and Lin, Yutong and Cao, Yue and Hu, Han and Wei, Yixuan and Zhang, Zheng and Lin, Stephen and Guo, Baining},
  booktitle={2021 IEEE/CVF International Conference on Computer Vision (ICCV)}, 
  title={{Swin Transformer: Hierarchical Vision Transformer using Shifted Windows}}, 
  year={2021},
  volume={},
  number={},
  pages={9992-10002},
  doi={10.1109/ICCV48922.2021.00986}
}

@INPROCEEDINGS{lam,
  author={Gu, Jinjin and Dong, Chao},
  booktitle={2021 IEEE/CVF Conference on Computer Vision and Pattern Recognition (CVPR)}, 
  title={{Interpreting Super-Resolution Networks with Local Attribution Maps}}, 
  year={2021},
  volume={},
  number={},
  pages={9195-9204},
  doi={10.1109/CVPR46437.2021.00908}}

@INPROCEEDINGS{swinir,
  author={Liang, Jingyun and Cao, Jiezhang and Sun, Guolei and Zhang, Kai and Van Gool, Luc and Timofte, Radu},
  booktitle={2021 IEEE/CVF International Conference on Computer Vision Workshops (ICCVW)}, 
  title={{SwinIR: Image Restoration Using Swin Transformer}}, 
  year={2021},
  volume={},
  number={},
  pages={1833-1844},
  doi={10.1109/ICCVW54120.2021.00210}
}

@INPROCEEDINGS{hat,
  author={Chen, Xiangyu and Wang, Xintao and Zhou, Jiantao and Qiao, Yu and Dong, Chao},
  booktitle={2023 IEEE/CVF Conference on Computer Vision and Pattern Recognition (CVPR)}, 
  title={{Activating More Pixels in Image Super-Resolution Transformer}}, 
  year={2023},
  volume={},
  number={},
  pages={22367-22377},
  doi={10.1109/CVPR52729.2023.02142}}

@INPROCEEDINGS{srformer,
  author={Zhou, Yupeng and Li, Zhen and Guo, Chun-Le and Bai, Song and Cheng, Ming-Ming and Hou, Qibin},
  booktitle={2023 IEEE/CVF International Conference on Computer Vision (ICCV)}, 
  title={{SRFormer: Permuted Self-Attention for Single Image Super-Resolution}}, 
  year={2023},
  volume={},
  number={},
  pages={12734-12745},
  doi={10.1109/ICCV51070.2023.01174}}

@inproceedings{mambair,
  title={{MambaIR: A Simple Baseline for Image Restoration with State-Space Model}},
  author={Guo, Hang and Li, Jinmin and Dai, Tao and Ouyang, Zhihao and Ren, Xudong and Xia, Shu-Tao},
  booktitle={European Conference on Computer Vision (ECCV)},
  pages={222--241},
  year={2024},
  organization={Springer},
  doi={10.1007/978-3-031-72649-1_13}
}

@INPROCEEDINGS{mambairv2,
  author={Guo, Hang and Guo, Yong and Zha, Yaohua and Zhang, Yulun and Li, Wenbo and Dai, Tao and Xia, Shu-Tao and Li, Yawei},
  booktitle={2025 IEEE/CVF Conference on Computer Vision and Pattern Recognition (CVPR)}, 
  title={{MambaIRv2: Attentive State Space Restoration}}, 
  year={2025},
  volume={},
  number={},
  pages={28124-28133},
  doi={10.1109/CVPR52734.2025.02619}}

@inproceedings{cat,
  title={{Cross Aggregation Transformer for Image Restoration}},
  author={Chen, Zheng and Zhang, Yulun and Gu, Jinjin and Zhang, Yongbing and Kong, Linghe and Yuan, Xin},
  booktitle={Advances in Neural Information Processing Systems (NeurIPS)},
  year={2022}
}

@inproceedings{art,
  title={{Accurate Image Restoration with Attention Retractable Transformer}},
  author={Zhang, Jiale and Zhang, Yulun and Gu, Jinjin and Zhang, Yongbing and Kong, Linghe and Yuan, Xin},
  booktitle={International Conference on Learning Representations (ICLR)},
  year={2023}
}

@inproceedings{ipg,
  title={{Image Processing GNN: Breaking Rigidity in Super-Resolution}},
  author={Tian, Yuchuan and Chen, Hanting and Xu, Chao and Wang, Yunhe},
  booktitle={2024 IEEE/CVF Conference on Computer Vision and Pattern Recognition (CVPR)},
  pages={24108--24117},
  year={2024}
}

@inproceedings{atd,
  title={{Transcending the Limit of Local Window: Advanced Super-Resolution Transformer with Adaptive Token Dictionary}},
  author={Zhang, Leheng and Li, Yawei and Zhou, Xingyu and Zhao, Xiaorui and Gu, Shuhang},
  booktitle={2024 IEEE/CVF Conference on Computer Vision and Pattern Recognition (CVPR)},
  pages={2856--2865},
  year={2024}
}

@inproceedings{pft,
  title={{Progressive Focused Transformer for Single Image Super-Resolution}},
  author={Long, Wei and Zhou, Xingyu and Zhang, Leheng and Gu, Shuhang},
  booktitle={2025 IEEE/CVF Conference on Computer Vision and Pattern Recognition (CVPR)},
  pages={2279--2288},
  year={2025}
}

@inproceedings{ucan,
  title={{UCAN: Unified Convolutional Attention Network for Expansive Receptive Fields in Lightweight Super-Resolution}},
  author={Tan, Cao Thien and Trang, Phan Thi Thu and Duc, Do Nghiem and Anh, Ho Ngoc and Zhuang, Hanyang and Dung, Nguyen Duc},
  booktitle={2026 IEEE/CVF Conference on Computer Vision and Pattern Recognition (CVPR)},
  pages={23409--23418},
  year={2026}
}

@inproceedings{mamba,
  title={{Mamba: Linear-Time Sequence Modeling with Selective State Spaces}},
  author={Albert Gu and Tri Dao},
  booktitle={First Conference on Language Modeling (COLM)},
  year={2024},
  doi={10.48550/arXiv.2312.00752}}

@INPROCEEDINGS{pixelshuffle,
  author={Shi, Wenzhe and Caballero, Jose and Huszár, Ferenc and Totz, Johannes and Aitken, Andrew P. and Bishop, Rob and Rueckert, Daniel and Wang, Zehan},
  booktitle={2016 IEEE Conference on Computer Vision and Pattern Recognition (CVPR)}, 
  title={{Real-Time Single Image and Video Super-Resolution Using an Efficient Sub-Pixel Convolutional Neural Network}}, 
  year={2016},
  doi={10.1109/CVPR.2016.207}}

@inproceedings{flashattention,
  title={{FlashAttention: Fast and Memory-Efficient Exact Attention with IO-Awareness}},
  author={Dao, Tri and Fu, Daniel Y. and Ermon, Stefano and Rudra, Atri and R{\'e}, Christopher},
  booktitle={Advances in Neural Information Processing Systems (NeurIPS)},
  year={2022},
  doi={10.48550/arXiv.2205.14135}}

@inproceedings{iet,
  author={Meng, Chunyu and Long, Wei and Gu, Shuhang},
  title={{From Local Windows to Adaptive Candidates via Individualized Exploratory: Rethinking Attention for Image Super-Resolution}},
  booktitle={European Conference on Computer Vision (ECCV)},
  year={2026}}

@misc{muon,
  author={Jordan, Keller and Jin, Yuchen and Boza, Vlado and You, Jiacheng and Cesista, Franz and Newhouse, Laker and Bernstein, Jeremy},
  title={{Muon: An Optimizer for Hidden Layers in Neural Networks}},
  year={2024},
}

@inproceedings{adam,
  author={Kingma, Diederik P. and Ba, Jimmy},
  title={{Adam: A Method for Stochastic Optimization}},
  booktitle={International Conference on Learning Representations (ICLR)},
  year={2015},
}

@inproceedings{mixed,
  author={Micikevicius, Paulius and Narang, Sharan and Alben, Jonah and Diamos, Gregory and Elsen, Erich and Garcia, David and Ginsburg, Boris and Houston, Michael and Kuchaiev, Oleksii and Venkatesh, Ganesh and Wu, Hao},
  title={{Mixed Precision Training}},
  booktitle={International Conference on Learning Representations (ICLR)},
  year={2018},
}
\bibliographystyle{iclr2027_conference}

\clearpage
\appendix
\numberwithin{equation}{section}
\section{Mathematical Properties of Graphene}\label{app:formal}

This appendix establishes properties of the Ising Refinement Block (IRB) and
stage-history attention (HistAttn). The analysis concerns their mathematical
formulations in exact real arithmetic. A companion Lean development checks
the coupling algebra, single-coordinate mean-field stationarity,
magnetization and gate bounds, conditional fixed-point convergence, and
history aggregation and fusion. The finite Gibbs-to-variational reduction is
derived explicitly below; the formal coverage is specified in
\Cref{app:formal-coverage}.

For one IRB, the token count is $n=W^2$, the field is
$h\in\mathbb{R}^{n}$, and the coupling matrix is
$J\in\mathbb{R}^{n\times n}$. The quantities $h$, $J$, and $\beta>0$
remain fixed during its $K_{\mathrm{mf}}$ synchronous updates. The coupling
construction considered here is the signed, symmetric configuration with
row-wise top-$k$ sparsification disabled. The notation $\|J\|_2$ denotes
the induced Euclidean operator norm. The convergence result concerns
repeated updates of this fixed-window map.

\subsection{Logit Lifting and the Final Coupling Matrix}\label{app:formal-couplings}

Let $d=\alpha_{\mathrm{psa}}^2>0$ and let $\pi(j)$ identify the reduced key
cell associated with fine token $j$. Each key-cell fiber contains exactly
$d$ fine tokens. For a fixed query $i$ and attention head $a$, the lifting
rule is
\begin{equation}\label{eq:app-lift}
\bar L^{(a)}_{ij}=\frac{L^{(a)}_{i,\pi(j)}}{d}.
\end{equation}

\begin{proposition}[Replication compensation]\label[proposition]{prop:app-lift}
For every reduced key cell $u$,
\begin{equation}
\sum_{j:\,\pi(j)=u}\bar L^{(a)}_{ij}=L^{(a)}_{iu}.
\end{equation}
\end{proposition}
\begin{proof}
Every term in this sum equals $L^{(a)}_{iu}/d$, and there are $d$ terms.
The sum is therefore $d(L^{(a)}_{iu}/d)=L^{(a)}_{iu}$. In Graphene,
$\alpha_{\mathrm{psa}}=2$, giving four copies scaled by $1/4$.
This identity applies before subsequent coupling transformations.
\end{proof}

Let $\tilde J$ denote the lifted, head-averaged logits after setting
invalid masked pairs to zero. Distinguishing the successive matrices gives
\begin{align}
\hat J_{ij}
  &=\tilde J_{ij}-\frac{1}{n}\sum_{k=1}^{n}\tilde J_{ik},
  \label{eq:app-center}\\
J^{\mathrm{sym}}
  &=\tfrac12(\hat J+\hat J^\top),\qquad
J_{ij}=\gamma J^{\mathrm{sym}}_{ij}\,\mathbf{1}[i\ne j],
  \quad\gamma>0.\label{eq:app-final-j}
\end{align}

\begin{proposition}[Coupling invariants]\label[proposition]{prop:app-couplings}
The centered intermediate satisfies $\sum_j\hat J_{ij}=0$ for each row.
The final matrix satisfies $J=J^\top$ and $J_{ii}=0$. Removing the diagonal
and symmetrizing commute, and a global scalar factor can be moved through
these two operations.
\end{proposition}
\begin{proof}
Summing \Cref{eq:app-center} over $j$ subtracts the same row sum that is
added. Symmetrization gives
$J^{\mathrm{sym}}_{ij}=J^{\mathrm{sym}}_{ji}$. Multiplication by the
symmetric indicator $\mathbf{1}[i\ne j]$ preserves this equality and
sets each diagonal entry to zero. For $i\ne j$, both orders of
symmetrization and diagonal removal give $(\hat J_{ij}+\hat J_{ji})/2$;
for $i=j$, both give zero. Scalar multiplication distributes through the
average and preserves the zero diagonal.
\end{proof}

\begin{remark}[The zero-row-sum property belongs to the intermediate]
The final row sums need not vanish. An example that respects Graphene's
actual two-by-two lifting uses a $4\times4$ window and four reduced key
cells. Set $L_{iu}=0$ when $u=\pi(i)$ and $L_{iu}=4$ otherwise, with
identical logits across heads and no mask. The lifted entries are zero
within each key-cell group and one across groups. Each row has mean
$3/4$, so the centered matrix is already symmetric, with entries $-3/4$
within a group and $1/4$ across groups. For $\gamma=1$, removing the
diagonal gives the first final row the sum
\begin{equation}
3\left(-\frac34\right)+12\left(\frac14\right)=\frac34.
\end{equation}
This example is also checked in Lean. The construction therefore gives a
symmetric matrix with a zero diagonal, without imposing a zero final row
sum. Constant input rows give $\hat J=0$ and $J=0$, so centering permits
signed deviations without requiring both signs for every input.
\end{remark}

\subsection{From the Ising Model to Mean-Field Stationarity}\label{app:formal-ising}

Consider spins $\xi_i\in\{-1,+1\}$ and the symmetric, diagonal-free
coupling matrix in \Cref{eq:app-final-j}. The Ising energy and Gibbs
distribution are
\begin{equation}
E(\boldsymbol\xi)
  =-\sum_i h_i\xi_i-\frac12\sum_{i,j}J_{ij}\xi_i\xi_j,
\qquad
p(\boldsymbol\xi)=Z^{-1}\exp[-\beta E(\boldsymbol\xi)].
\end{equation}
For a factorized distribution $q(\boldsymbol\xi)=\prod_i q_i(\xi_i)$,
write $q_i(+1)=(1+m_i)/2$ and $q_i(-1)=(1-m_i)/2$.
For interior means $m_i\in(-1,1)$, minimizing
$\mathrm{KL}(q\|p)$ is equivalent to minimizing, up to the constant
$\log Z$, the dimensionless variational objective
\begin{align}
\Phi(m)
  &=-\beta\sum_i h_i m_i
    -\frac{\beta}{2}\sum_{i,j}J_{ij}m_i m_j
    +\sum_i\psi(m_i),\label{eq:app-free-energy}\\
\psi(t)
  &=\frac{1+t}{2}\log\frac{1+t}{2}
    +\frac{1-t}{2}\log\frac{1-t}{2}.\label{eq:app-entropy}
\end{align}
Indeed, $\mathrm{KL}(q\|p)=\mathbb E_q[\log q]+
\beta\mathbb E_q[E]+\log Z$. Factorization separates the entropy
terms and gives $\mathbb E_q[\xi_i\xi_j]=m_i m_j$ for $i\ne j$;
the diagonal terms vanish because $J_{ii}=0$.

\begin{proposition}[Mean-field self-consistency]\label[proposition]{prop:app-stationary}
An interior stationary point of \Cref{eq:app-free-energy} satisfies
\begin{equation}\label{eq:app-self-consistency}
m_i=\tanh\!\left(\beta\left(h_i+\sum_jJ_{ij}m_j\right)\right)
\quad\text{for every }i.
\end{equation}
Conversely, an interior solution of \Cref{eq:app-self-consistency} is a
stationary point of \Cref{eq:app-free-energy}.
\end{proposition}
\begin{proof}
Differentiating the pair term yields both a row contribution and a column
contribution. Symmetry gives
\begin{equation}
\frac12\left(\sum_jJ_{ij}m_j+\sum_jJ_{ji}m_j\right)
  =\sum_jJ_{ij}m_j.
\end{equation}
The entropy derivative is
\begin{equation}
\psi'(t)=\frac12\left(\log\frac{1+t}{2}-\log\frac{1-t}{2}\right).
\end{equation}
Holding all other means fixed, define the effective field
$x_i=\beta(h_i+\sum_jJ_{ij}m_j)$. The condition $J_{ii}=0$ makes
$x_i$ independent of the coordinate being varied. The objective as a
function of that coordinate is $-x_i t+\psi(t)$ plus a constant.
Its stationary equation is
\begin{equation}
\frac12\log\frac{1+t}{1-t}=x_i.
\end{equation}
Solving this equation gives
\begin{equation}
t=\frac{e^{2x_i}-1}{e^{2x_i}+1}
  =\frac{e^{x_i}-e^{-x_i}}{e^{x_i}+e^{-x_i}}
  =\tanh(x_i).
\end{equation}
All steps are reversible for $t\in(-1,1)$. The Lean development checks
the single-coordinate derivative and the equivalence between a zero
derivative and $t=\tanh(x_i)$, together with the symmetric field identity.
\end{proof}

The same expression is the mean of a single binary spin with probabilities
proportional to $e^{x_i}$ and $e^{-x_i}$. Replacing neighboring spins by
their current means motivates Graphene's synchronous update
$m^{(t+1)}=\tanh(\beta(h+Jm^{(t)}))$.
A finite unroll evaluates this update rule; stationarity of the
variational objective is a property of a self-consistent solution.

\subsection{Magnetization Bounds and the Residual Gate}\label{app:formal-gate}

\begin{proposition}[Bounded magnetizations]\label[proposition]{prop:app-bounds}
For finite real fields, couplings, and inputs, every coordinate produced
by a mean-field update lies in $(-1,1)$. The same holds for the
$\tanh$-initialized magnetizations. Consequently, every finite IRB
unroll satisfies $m_i^{(t)}\in[-1,1]$.
\end{proposition}
\begin{proof}
For every real $x$,
$-1<\tanh x<1$. Apply this fact to the affine initialization and to
each coordinate of each update. No contraction assumption is needed
for this bound.
\end{proof}

\begin{proposition}[Gate range and correction magnitude]\label[proposition]{prop:app-gate}
Let $g_i=\sigma(\kappa m_i)$, where
$\sigma(x)=(1+e^{-x})^{-1}$, $\kappa\ge0$, and $m_i\in[-1,1]$.
Then
\begin{equation}
\sigma(-\kappa)\le g_i\le\sigma(\kappa),\qquad
\sigma(\kappa\cdot0)=\frac12.
\end{equation}
For the residual update $y'_i=y_i+g_iR(y_i)$ in a real normed feature
space,
\begin{equation}\label{eq:app-residual-norm}
\|y'_i-y_i\|=g_i\|R(y_i)\|.
\end{equation}
\end{proposition}
\begin{proof}
The bound $-\kappa\le\kappa m_i\le\kappa$ and monotonicity of the
sigmoid give the gate interval. Substitution gives $\sigma(0)=1/2$.
The residual difference is $g_iR(y_i)$; absolute homogeneity of the norm
and $g_i>0$ give \Cref{eq:app-residual-norm}.
\end{proof}

For $\kappa=1$, the gate interval is approximately $[0.269,0.731]$.
Thus, neutral magnetization gives half of the learned residual.
The identity path preserves the input feature, while proximity of the
output to that feature depends on the magnitude of the gated residual.
In particular, $R(y_i)=0$ gives $y'_i=y_i$ for any magnetization.

\subsection{A Sufficient Condition for Fixed-Window Convergence}\label{app:formal-stability}

Define $F:\mathbb R^n\to\mathbb R^n$ by
\begin{equation}
F(m)=\tanh\bigl(\beta(h+Jm)\bigr),
\end{equation}
with componentwise $\tanh$ and fixed $h$, $J$, and $\beta\ge0$.

\begin{lemma}[Euclidean Lipschitz bound]\label[lemma]{lem:app-lipschitz}
For any $m,m'\in\mathbb R^n$,
\begin{equation}\label{eq:app-lipschitz}
\|F(m)-F(m')\|_2
  \le\beta\|J\|_2\|m-m'\|_2.
\end{equation}
\end{lemma}
\begin{proof}
The scalar derivative satisfies
$\tanh'(x)=1/\cosh^2(x)\in(0,1]$. The mean value theorem gives
$|\tanh(x)-\tanh(y)|\le|x-y|$. Squaring the coordinatewise
inequalities, summing them, and taking square roots proves that
componentwise $\tanh$ is $1$-Lipschitz in the Euclidean norm. Therefore,
\begin{align}
\|F(m)-F(m')\|_2
&\le\|\beta(h+Jm)-\beta(h+Jm')\|_2\\
&=\beta\|J(m-m')\|_2
 \le\beta\|J\|_2\|m-m'\|_2.
\end{align}
The final inequality is the defining property of the induced operator
norm. The formalization uses Euclidean token space explicitly.
\end{proof}

\begin{theorem}[Conditional convergence]\label{thm:app-convergence}
If $q=\beta\|J\|_2<1$, then $F$ has a unique fixed point
$m^\star\in\mathbb R^n$. For every initialization $m^{(0)}$, the
synchronous iteration converges to $m^\star$. Moreover,
\begin{equation}\label{eq:app-error}
\|m^{(K)}-m^\star\|_2
  \le q^K\|m^{(0)}-m^\star\|_2
\quad\text{for every integer }K\ge0.
\end{equation}
\end{theorem}
\begin{proof}
By \Cref{lem:app-lipschitz}, $F$ is a contraction with constant $q$ on
the complete space $\mathbb R^n$. The Banach fixed-point theorem gives
existence, uniqueness, and convergence from every initialization.
Applying the Lipschitz inequality repeatedly to $m^{(0)}$ and the fixed
point gives \Cref{eq:app-error}. Since $F(m^\star)=m^\star$,
\Cref{prop:app-bounds} also places every coordinate of $m^\star$ in
$(-1,1)$.
\end{proof}

\Cref{thm:app-convergence} supplies a sufficient condition for the
fixed-window iteration. The finite IRB uses a prescribed
$K_{\mathrm{mf}}$ and does not enforce $\beta\|J\|_2<1$.
The boundedness result in \Cref{prop:app-bounds} holds independently of
this condition. The theorem applies separately to each fixed-window
map; it does not combine successive IRBs into iterations of one common
map.

\subsection{History Aggregation and Residual Fusion}\label{app:formal-history}

Fix a HistAttn insertion $s$ and spatial token position $p$, and suppress
$p$ for clarity. For finite compatibility scores $e_0,\ldots,e_s$,
define
\begin{equation}
w_r=\frac{e^{e_r}}{\sum_{\ell=0}^{s}e^{e_\ell}},\qquad
c=\sum_{r=0}^{s}w_r z^{(r)},\qquad
\hat z=z^{(s)}+\lambda_s(c-z^{(s)}).
\end{equation}

\begin{proposition}[Convex selection and affine fusion]\label[proposition]{prop:app-history}
The weights satisfy $w_r>0$ and $\sum_r w_r=1$, so $c$ belongs to the
convex hull of the stored features. The fused feature satisfies
\begin{equation}
\hat z=(1-\lambda_s)z^{(s)}+\lambda_s c
      =\sum_{r=0}^{s}\tilde w_r z^{(r)},
\end{equation}
where
\begin{equation}\label{eq:app-effective-weights}
\tilde w_r=\lambda_s w_r\quad(r<s),\qquad
\tilde w_s=1-\lambda_s(1-w_s),\qquad
\sum_{r=0}^{s}\tilde w_r=1.
\end{equation}
In particular, $\lambda_s=0$ gives $\hat z=z^{(s)}$ and
$\lambda_s=1$ gives $\hat z=c$.
\end{proposition}
\begin{proof}
Every exponential is positive, and the denominator is a nonempty sum
of positive terms. Summing the normalized weights gives one, establishing
convex-hull membership of $c$. Expanding the residual fusion gives
$(1-\lambda_s)z^{(s)}+\lambda_s\sum_r w_r z^{(r)}$.
Collecting the coefficient of the current feature yields
\Cref{eq:app-effective-weights}; its coefficient sum is
$1-\lambda_s+\lambda_s\sum_r w_r=1$.
The endpoint identities follow by substitution.
\end{proof}

The learned scalar $\lambda_s\in\mathbb R$ is unconstrained.
The sufficient conditions
\begin{equation}
\lambda_s\ge0,\qquad \lambda_s(1-w_s)\le1
\end{equation}
make all effective coefficients nonnegative. These conditions can hold
for some $\lambda_s>1$, since the current feature is itself part of the
selected history. In general, however, the complete fusion can leave the
convex hull. For example, scalar stored features $z^{(0)}=1$ and
$z^{(1)}=0$, uniform softmax weights, and $\lambda_s=3$ give
$c=1/2$ and $\hat z=3/2$. This exact example and the sufficient
coefficient condition are checked in Lean.

\clearpage
\subsection{Formal Verification and Reproducibility}\label{app:formal-coverage}

The companion project uses Lean 4.33.1~\cite{graphenelean4} and the
matching mathlib release~\cite{graphenemathlib4}. Its six modules contain
56 checked lemmas and theorems, including supporting identities.
\Cref{tab:app-proof-coverage} summarizes their scope. Every named theorem
passes a transitive axiom audit: dependencies are limited to Lean's
standard propositional extensionality, classical choice, and quotient
soundness axioms. The development contains no admitted proof, custom
unproved axiom, or native computation oracle.

\begin{table}[!ht]
\caption{Scope of the companion Lean development. The finite probability-model
reduction leading to \Cref{eq:app-free-energy} is supplied as a conventional
derivation; the formalized Ising statements cover the indicated algebra and
single-coordinate calculus.}
\label{tab:app-proof-coverage}
\centering
\small
\renewcommand{\arraystretch}{1.12}
\begin{tabular}{@{}p{0.22\linewidth}p{0.73\linewidth}@{}}
\toprule
Lean module & Checked properties \\
\midrule
\texttt{Lifting} & Group-sum preservation for a nonempty replication group and a general key-cell fiber. \\
\texttt{Couplings} & Centered row sums, symmetry, zero diagonal, commuting transformations, constant-row degeneracy, and the admissible lifted counterexample. \\
\texttt{Ising} & Binary-state probabilities and mean; single-coordinate entropy and free-energy derivatives; stationarity equivalence; symmetric field identity. \\
\texttt{Bounds} & Magnetization and gate bounds, the neutral gate, and gated residual norm identities. \\
\texttt{Stability} & Euclidean Lipschitz bounds, bounded finite iterates, conditional unique fixed point and convergence, and the finite-iteration error bound. \\
\texttt{History} & Positive normalized weights, convex-hull membership, effective fusion coefficients, endpoint identities, and an extrapolation example. \\
\bottomrule
\end{tabular}
\end{table}

The proof project pins the toolchain and dependency revisions and provides
a build-and-audit script. The mathematical definitions were compared
with the archived implementation used for the coupling diagnostics,
the matching evaluation implementation, and the current training source.
Diagnostic execution of the original IRB and HistAttn class bodies
agrees with the stated formulas on the recorded inputs. These execution
checks document source correspondence; the Lean results concern exact
real arithmetic and do not certify floating-point execution, optimization
dynamics, or reconstruction quality.

\clearpage
\section{Architecture and Evaluation Details}\label{app:implementation}

\subsection{Full and Lightweight Configurations}

\Cref{tab:architectures} specifies the two architecture sizes. Both use six attention heads per block, a $5\times5$ depth-wise convolution in ConvFFN, a one-convolution residual connection around each stage, and PSA reduction factor two. Each scale uses a separate checkpoint and reconstruction head. Graphene-light reduces the feature width and stage count and places one IRB at the end of each stage.

\begin{table}[htbp]
\caption{Architecture and refinement settings. Scalars described as fixed are not learned. The lightweight residual scalar is included in the effective residual map $R$ used in the main formulation.}
\label{tab:architectures}
\centering
\begin{tabular}{lcc}
\toprule
Setting & Graphene & Graphene-light \\
\midrule
Stages & 6 & 4 \\
Blocks per stage & 6 & 6 \\
Feature channels & 180 & 60 \\
Attention heads per block & 6 & 6 \\
Window side length & 32 & 16 \\
PSA reduction factor & 2 & 2 \\
ConvFFN hidden-width ratio & 2 & 1 \\
ConvFFN depth-wise kernel & $5\times5$ & $5\times5$ \\
IRB placement & Every block & Last block per stage \\
Number of IRBs & 36 & 4 \\
Mean-field updates per IRB & 3 & 3 \\
Residual MLP hidden-width ratio & 1 & 1 \\
$\beta$, $\gamma$, $\kappa$ & $1,1,1$ (fixed) & $1,1,1$ (fixed) \\
Signed, symmetric couplings & Enabled & Enabled \\
Row-wise top-$k$ pruning & Disabled & Disabled \\
Additional IRB residual scalar & None & Learned; initialized to 0 \\
HistAttn insertions & 5 & 3 \\
History fusion $\lambda_s$ & Learned; initialized to 0 & Learned; initialized to 0 \\
Recent-stage bias $\rho$ & 6 (fixed) & 6 (fixed) \\
Reconstruction head & PixelShuffle & Direct PixelShuffle \\
\bottomrule
\end{tabular}
\end{table}

The IRB residual MLP consists of LayerNorm, a linear map to the hidden width, GELU, and a linear map back to the feature width. The additional lightweight scalar multiplies this residual before gated addition. Its sign and magnitude are learned independently of the bounded sigmoid gate. HistAttn stores the shallow feature and successive stage outputs; the current output is included in its selection set. Zero initialization of $\lambda_s$ makes each history insertion an identity map at initialization.

\subsection{PSA Geometry and Coupling Construction}

For window features $X\in\mathbb R^{W^2\times C_{\mathrm{feat}}}$, PSA~\cite{srformer} retains $W^2$ full-width query tokens. The key/value projections have $C_{\mathrm{feat}}/\alpha_{\mathrm{psa}}^2$ channels before permutation. Concatenating each $\alpha_{\mathrm{psa}}\times\alpha_{\mathrm{psa}}$ spatial group yields $W^2/\alpha_{\mathrm{psa}}^2$ full-width key/value tokens. Each head therefore operates on logits of shape $W^2\times(W^2/\alpha_{\mathrm{psa}}^2)$, with head dimension $d_h=C_{\mathrm{feat}}/H$.

IRB maps each reduced-key entry back to its corresponding fine tokens and divides by the replication count. Head averaging, mask-sentinel removal, row-centering, symmetrization, and diagonal removal then produce one coupling matrix per window. Sentinel removal occurs before centering. The final coupling matrix can contain nonzero entries at positions that were masked during attention. Such positions occur only in the border-wrapping windows of shifted-window blocks. There $\tilde J_{ij}=0$, so \Cref{eq:coupling} gives $J_{ij}=-\tfrac{\gamma}{2}(\mu_i+\mu_j)$, with $\mu_i$ the row mean of $\tilde J$: a sum of per-token terms with no query--key term between $i$ and $j$. Because $R(y_i)$ depends on token $i$ alone, such a coupling affects only the scalar gate $g_i$. The properties of this sequence are established in Appendix~\ref{app:formal}.

\subsection{Reconstruction Metrics and Image Inference}

The benchmark comparison uses reported baseline PSNR/SSIM values, while Graphene reconstructions are evaluated through the same stated Y-channel metric convention. RGB predictions are clamped to $[0,1]$, multiplied by 255, rounded, and saved as 8-bit PNGs. The MATLAB evaluation converts RGB images with \texttt{rgb2ycbcr} applied to \texttt{im2double}, retains floating-point Y values, and multiplies Y by 255. A border of $r$ pixels is removed on each side for scale factor $r$. PSNR uses the mean squared error on the remaining pixels with dynamic range 255; SSIM uses the corresponding Y images. Dataset scores average the per-image values.

Classical and lightweight Graphene inference uses FP32, EMA checkpoints, a single forward pass per image, and no geometric self-ensemble. Right and bottom flip-concatenation padding makes the LR dimensions compatible with the attention window. Predictions are cropped back to the scaled original extent. These full-image benchmark measurements are separate from the fixed-crop continuation diagnostic in Section~\ref{app:paired}. The real-world examples use the separate released real-world checkpoint with BF16 eager inference and reflection padding; no PSNR/SSIM or runtime comparison is reported for those unpaired images.

\makeatletter\global\@fptop=0pt\relax\makeatother
\clearpage
\makeatletter\global\@fptop=0pt plus 1fil\relax\makeatother
\section{Additional Ablations and Diagnostics}\label{app:additional-experiments}

\subsection{Mean-Field Unroll and Coupling Design}\label{app:unroll}

\Cref{tab:ablation} extends the main component analysis. All variants use the six-stage, 180-channel backbone, both refinement modules, and window side length 32. Training uses DIV2K for $10{,}000$ iterations, with evaluation on Urban100 at $\times2$. Increasing $K_{\mathrm{mf}}$ from one to three improves PSNR; longer unrolls do not improve PSNR after the same training duration. Three updates give the highest PSNR among the tested lengths with fewer recurrent updates than the ten- and one-hundred-update variants.

\begin{table}[htbp]
\caption{Mean-field and coupling ablations on Urban100 at $\times2$.}
\label{tab:ablation}
\centering
\begin{tabular}{ccccc}
\toprule
$K_{\mathrm{mf}}$ & Signed $J$ & Symmetric $J$ & PSNR & SSIM \\
\midrule
1 & Enabled & Enabled & 30.42 & 0.915 \\
2 & Enabled & Enabled & 30.84 & 0.915 \\
3 & Enabled & Enabled & \textbf{30.93} & \textbf{0.917} \\
10 & Enabled & Enabled & 30.85 & 0.915 \\
100 & Enabled & Enabled & 30.85 & 0.915 \\
3 & Disabled & Enabled & 30.90 & 0.916 \\
3 & Enabled & Disabled & 30.86 & 0.916 \\
\bottomrule
\end{tabular}
\end{table}

In the nonnegative variant, negative entries are clipped to zero after row-centering and before symmetrization. Clipping changes both the permitted signs and the aggregate coupling magnitude. The coupling-design comparison retains the same residual MLP architecture. By contrast, adding IRB in the component comparison also adds the residual MLP and its parameters. The coupling ablation therefore compares interaction rules, while the component ablation measures the PSNR change from the entire IRB.

\noindent\textbf{Gate controls.} \Cref{tab:component_ablation}b retains the residual MLP, PSA, and HistAttn. The constant gate uses $\tilde y_i=y_i+0.5R(y_i)$ and removes both scalar projections. With $J=0$, the unary-only gate is $g_i=\sigma(\kappa\tanh(\beta h_i))$; its unused initial-magnetization projection and pairwise recurrence are omitted. Both retain the same residual MLP. The controls train from scratch for $10{,}000$ iterations with matching shared-parameter initialization, minibatches, and optimization settings.

\noindent\textbf{History controls.} \Cref{tab:history_selection} retains full IRB. The shallow skip sets $c_p^{(s)}=z_p^{(0)}$. Static selection uses learned logits $a_r^{(s)}$ and weights $\bar w_r^{(s)}=\exp(a_r^{(s)})/\sum_{\ell=0}^{s}\exp(a_\ell^{(s)})$, shared across images and tokens, to form $c_p^{(s)}=\sum_{r=0}^{s}\bar w_r^{(s)}z_p^{(r)}$. Static selection and HistAttn use identical history entries; the static and shallow controls omit feature scoring and its LayerNorm. All enabled variants retain $\hat z_p^{(s)}=z_p^{(s)}+\lambda_s(c_p^{(s)}-z_p^{(s)})$ at the same insertions, with the same initialization of $\lambda_s$.

\subsection{Variation Across Initializations}\label{app:seed-variation}

\Cref{tab:statistical_significance} reports all six runs with IRB and HistAttn enabled under the $10{,}000$-iteration DIV2K protocol. The mean is $30.92$ dB PSNR with sample SD $0.08$ dB, and $0.915$ SSIM with sample SD $0.001$. These statistics describe repeatability across initializations for the evaluated configuration. They are distinct from uncertainty in a paired difference between two architectures.

\begin{table}[htbp]
\caption{Urban100 $\times2$ results across six independent training seeds. SD denotes the sample standard deviation.}
\label{tab:statistical_significance}
\centering
\setlength{\tabcolsep}{4pt}
\begin{tabular}{lrrrrrrrr}
\toprule
Metric & 4128 & 4129 & 4130 & 4127 & 4131 & 4132 & Mean & SD \\
\midrule
PSNR & 31.02 & 30.90 & 30.81 & 30.94 & 30.86 & 30.99 & 30.92 & 0.08 \\
SSIM & 0.916 & 0.915 & 0.913 & 0.915 & 0.914 & 0.916 & 0.915 & 0.001 \\
\bottomrule
\end{tabular}
\end{table}

\subsection{Transfer to HAT}\label{app:irb-transferability}

\Cref{tab:irb_transfer} compares HAT and HAT+IRB after $50{,}000$ training iterations. IRB improves all five $\times2$ PSNR values by $0.16$--$0.94$ dB. At $\times4$, it improves four of five datasets by up to $0.21$ dB and decreases Set5 by $0.02$ dB. The table also reports parameter counts and FLOPs for the IRB-augmented HAT models.

\begin{table}[htbp]
\caption{Transfer of IRB to HAT after $50{,}000$ training iterations. Dataset columns report PSNR (dB).}
\label{tab:irb_transfer}
\centering
\small
\begin{tabular}{clrrrrrrr}
\toprule
Scale & Model & Params & FLOPs & Set5 & Set14 & B100 & U100 & M109 \\
& & (M) & (G) & & & & & \\
\midrule
$\times2$ & HAT & 20.6 & 203.4 & 37.97 & 33.48 & 32.25 & 32.21 & 38.77 \\
$\times2$ & HAT+IRB & 22.8 & 226.1 & 38.16 & 33.82 & 32.41 & 33.15 & 39.36 \\
$\times4$ & HAT & 20.8 & 208.4 & 32.40 & 28.54 & 27.78 & 26.75 & 31.36 \\
$\times4$ & HAT+IRB & 22.9 & 231.0 & 32.38 & 28.55 & 27.86 & 26.96 & 31.55 \\
\bottomrule
\end{tabular}
\end{table}

\FloatBarrier
\subsection{Paired Continuation and Regional Metrics}\label{app:paired}

\noindent\textbf{Initialization and training protocol.} Both $\times4$ variants are initialized from the same pretrained Graphene model, distinct from the model used for the final benchmark comparisons. The unary-only variant sets $J=0$ in all 36 IRBs while retaining the unary field, sigmoid gate, residual MLP, PSA, and HistAttn. Both arms run $1{,}000$ optimizer steps on DF2K with RGB $\ell_1$ loss. Adam is reset in both arms with learning rate $10^{-5}$, $(\beta_1,\beta_2)=(0.9,0.999)$, $\epsilon=10^{-8}$, zero weight decay, and a constant schedule. Four microbatches of eight give effective batch size 32. LR training crops are $64\times64$ with matched flips and transposes. Both arms receive the same sampled tensors and stochastic seeds. The run uses FP32 without autocast or TF32 and evaluates ordinary weights without EMA.

\noindent\textbf{Evaluation regions.} Evaluation uses fixed $128\times128$ LR crops from all 100 Urban100 images and 16 synthetic cases. Scores use the central region after removing $32r$ HR pixels per side at scale factor $r$, excluding invalid pixels. For natural images, Canny edges are computed from HR luminance with smoothing $\sigma=r$ and thresholds 0.05 and 0.10. Boundary pixels lie within an $r$-iteration binary dilation of these edges. Interior pixels lie outside a $3r$-iteration dilation. Synthetic cases use known region boundaries. These masks define appearance-based regional diagnostics. Predictions are clipped to $[0,1]$ and converted to floating-point Y without PNG quantization. Regional PSNR uses that luminance MSE with unit dynamic range.

\noindent\textbf{Statistics.} Intact-minus-unary differences are calculated separately for each image. The reported natural-image intervals are percentile intervals from 10,000 image-level bootstrap resamples, with no multiplicity correction. Boundary-minus-interior contrasts use the same 98 images with valid interior pixels. The fixed final step is the comparison endpoint. Intervals describe image-level variation for one paired training seed and do not estimate variation across training seeds.

\begin{table}[htbp]
\caption{Trajectory of the paired $\times4$ continuation on Urban100 crops. Values are intact-minus-unary PSNR gaps (dB), with exploratory 95\% image-level intervals.}
\label{tab:paired-trajectory}
\centering
\small
\setlength{\tabcolsep}{4pt}
\begin{tabular}{rccc}
\toprule
Step & All pixels ($n=100$) & Boundary ($n=100$) & Boundary minus interior ($n=98$) \\
\midrule
0 & $0.0076\ [0.0020,0.0133]$ & $0.0061\ [-0.0024,0.0139]$ & $0.0035\ [-0.0079,0.0140]$ \\
250 & $0.0095\ [0.0044,0.0148]$ & $0.0080\ [0.0013,0.0146]$ & $0.0030\ [-0.0065,0.0118]$ \\
500 & $0.0093\ [0.0027,0.0169]$ & $0.0078\ [-0.0009,0.0168]$ & $0.0029\ [-0.0085,0.0142]$ \\
750 & $0.0105\ [0.0045,0.0172]$ & $0.0083\ [0.0003,0.0165]$ & $0.0050\ [-0.0054,0.0153]$ \\
1000 & $0.0163\ [0.0094,0.0243]$ & $0.0139\ [0.0055,0.0228]$ & $0.0023\ [-0.0087,0.0138]$ \\
\bottomrule
\end{tabular}
\end{table}

The all-pixel gap in \Cref{tab:paired-trajectory} increases from $0.0076$ to $0.0163$ dB ($+0.0087$ dB; image-bootstrap interval $[0.0033,0.0151]$ dB). Intact Graphene outperforms the unary-only variant on 67 of 100 final crops. Absolute PSNR decreases from $28.1408$ to $28.0426$ dB for intact Graphene and from $28.1332$ to $28.0262$ dB for the unary-only variant, indicating a relative advantage under reset-optimizer continuation. The final boundary-minus-interior interval spans zero. Synthetic mean all-pixel, boundary, and boundary-minus-interior gaps are $0.0732$, $0.0827$, and $0.0446$ dB, respectively, at the final step.

\clearpage
\section{Feature and Attribution Analyses}\label{app:features}

\subsection{Couplings Across Image Regions}

\Cref{fig:irb_history_analysis} examines the non-shifted IRB in Stage~6, Block~5 of the full $\times2$ model. The analysis covers $18{,}221$ non-overlapping $32\times32$ LR windows from Urban100, excluding windows that extend into image padding. Disjoint extreme-decile categories select $1{,}822$ windows each. Flat windows have low variation, edge windows have concentrated gradients, and texture windows have distributed high-frequency responses.

\noindent\textbf{Region selection.} For RGB inputs in $[0,1]$, grayscale intensity is $I=0.299R+0.587G+0.114B$. Let $g$ be the magnitude of horizontal and vertical backward differences within a window, with missing differences set to zero, and let $\ell$ be the absolute four-neighbor Laplacian, set to zero on the window border. Write $\bar g$, $g_{95}$, $v$, $\bar\ell$, and $s_\ell$ for the mean gradient magnitude, its 95th percentile, intensity variance, mean absolute Laplacian, and population SD of the absolute Laplacian. The normalized rank $P(a)\in[0,1]$ orders each statistic across all valid windows, with ties resolved by image name and window order. Define gradient sparsity $d=P(g_{95})-P(\bar g)$ and the scores
\[
\begin{aligned}
a_{\mathrm{flat}}&=-0.65P(\bar g)-0.35P(v),\\
a_{\mathrm{edge}}&=0.55P(g_{95})+0.55d+0.15P(v)-0.25P(\bar\ell),\\
a_{\mathrm{texture}}&=0.40P(\bar g)+0.35P(s_\ell)+0.25P(v)-0.20d.
\end{aligned}
\]
The highest-scoring windows are selected in flat, edge, then texture order, excluding previously selected windows. Within each category, the displayed representative minimizes the distance between its mean absolute coupling and the category median of per-window means; ties are resolved by image name and window index.

The category-averaged mean absolute couplings are $0.044486$, $0.281280$, and $0.349161$ for flat, edge, and texture windows, respectively. The corresponding edge-to-flat and texture-to-flat ratios are $6.3$ and $7.8$, calculated before rounding. The individual displayed windows have means $0.039$, $0.250$, and $0.318$. These statistics show that coupling magnitudes vary with image content. Positive and negative entries influence latent-spin alignment and opposition; regional reconstruction effects are measured separately in the paired continuation study.

\FloatBarrier
\subsection{History Selection and Effective Fusion}

The history analysis pools $19{,}831{,}808$ spatial tokens per insertion across Urban100, including positions introduced by window padding. Modules feeding Stages~3--4 mainly select the shallow representation $z^{(0)}$. The module feeding Stage~6 assigns $40.8\%$ to $z^{(5)}$ and $59.2\%$ to earlier features. These percentages describe the selected aggregate before multiplication by the residual-fusion strength.

The effective coefficients are $\tilde w_r=\lambda_s w_r$ for $r<s$ and $\tilde w_s=1-\lambda_s(1-w_s)$ for the current feature. \Cref{tab:history-effective} reports these quantities separately. At the final insertion, the mean effective coefficient is approximately $0.605$ for the current feature, and the earlier coefficients sum to approximately $0.395$. These are coefficient masses, not fractions of feature energy or causal influence. Their difference from the selection weights follows from the identity term in residual fusion.

Selection also varies across the collected tokens. The shallow-source weight entering Stage~2 has mean $0.593$ and token-level SD $0.179$; the current-source weight entering Stage~6 has mean $0.408$ and SD $0.106$. These pooled statistics establish nonconstant selection in the evaluated inputs. \Cref{tab:history_selection} compares HistAttn with a learned shallow skip and learned static weights at the same insertions.

\begin{table}[htbp]
\caption{Mean history-selection weights and effective fusion coefficients. The insertion immediately preceding the listed destination stage is evaluated. Intermediate history sources are omitted from the displayed columns.}
\label{tab:history-effective}
\centering
\begin{tabular}{rrrrrr}
\toprule
Destination & $w_0$ & $w_s$ & $\lambda_s$ & $\tilde w_0$ & $\tilde w_s$ \\
\midrule
2 & 0.593 & 0.407 & 0.500 & 0.297 & 0.703 \\
3 & 0.969 & 0.030 & 0.832 & 0.806 & 0.193 \\
4 & 0.988 & 0.009 & 0.815 & 0.805 & 0.192 \\
5 & 0.555 & 0.029 & 0.858 & 0.477 & 0.167 \\
6 & 0.178 & 0.408 & 0.667 & 0.119 & 0.605 \\
\bottomrule
\end{tabular}
\end{table}

\FloatBarrier
\subsection{Stage-Wise and Cross-Scale Activations}

The stage-wise activation figure in the main paper uses channel-mean maps on Urban100 img\_024. The PSA row records stage outputs from an independently trained SRFormer checkpoint. The IRB and HistAttn rows use Graphene with the other refinement module disabled at inference. The IRB row records stage outputs with HistAttn disabled; the HistAttn row records features after history fusion at Stages~1--5. Stage~6 has no subsequent history fusion. Each panel is divided by its own 99.5th percentile of absolute values and clipped to $[-1,1]$. Consequently, color reflects relative within-panel spatial organization, without a common amplitude scale across configurations.

\begin{figure}[htbp]
\centering
\includegraphics[width=\textwidth]{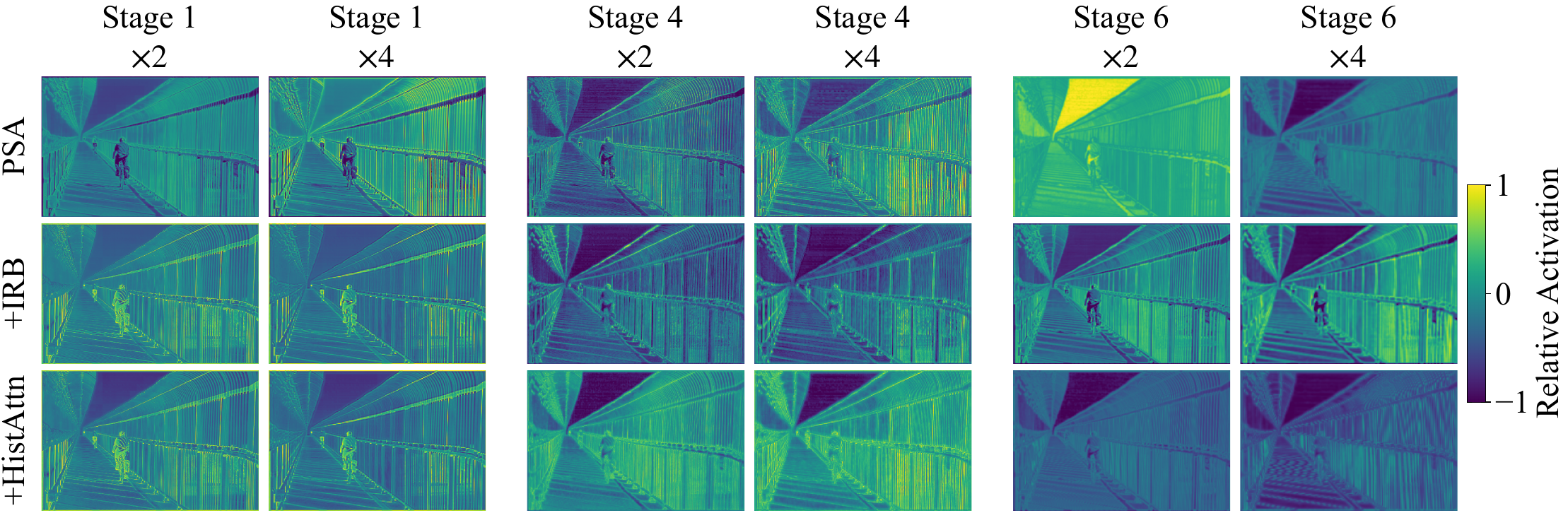}
\caption{Cross-scale stage activations on Urban100 img\_024. Each panel is independently normalized. IRB and HistAttn display positively corresponding spatial responses, while PSA retains recognizable structure with reversed polarity at Stage~6. Amplitude ratios are computed from the unnormalized maps.}
\label{fig:cross_scale_feature}
\end{figure}

\Cref{fig:cross_scale_feature} extends the illustration to independently trained $\times2$ and $\times4$ models. The IRB and HistAttn maps retain recognizable spatial organization across the displayed stages. PSA also retains scene structure at Stage~6, with reversed response polarity. At Stage~6, the $\times4$-to-$\times2$ activation-amplitude ratios are $0.639$ for IRB, $0.440$ for HistAttn, and $0.103$ for PSA. These ratios use mean absolute values of the underlying unnormalized channel-mean maps. They measure relative amplitude, whereas the independently normalized panels display relative spatial patterns.

\FloatBarrier
\subsection{Local Attribution Map Protocol}\label{app:attribution}

LAM~\cite{lam} is evaluated on Urban100 img\_045 at $\times4$. The LR context is a $128\times128$ crop with full-image origin $(128,0)$. The output region has full-HR origin $(800,90)$ and size $100\times44$; within the cropped prediction its origin is $(288,90)$. The Gaussian path uses eight steps, $\sigma=1.2$, and kernel size nine. The model-specific forward computation is differentiated through the same attribution procedure.

The differentiated objective sums the output spatial-gradient magnitudes over channels and pixels in the target region. Each magnitude is the Euclidean norm of horizontal and vertical forward differences. Path-integrated input derivatives are summed across RGB channels, converted to absolute values, and normalized by the spatial maximum to give a map $a$. The implementation computes $\mathrm{DI}=100[1-\mathrm{Gini}(a+10^{-7})]$, where the small offset stabilizes the Gini calculation. Larger DI indicates more spatially distributed attribution.

\Cref{tab:lam-di} reports DI values from the runs used to generate the displayed maps, rounded to integers in the main figure. MambaIRv2 uses stochastic routing: its DI is $18.6200$ in the displayed run and $18.7562$ in a seed-zero rerun, both rounding to 19. The analysis is specific to the selected image, crop, and attribution setup.

\begin{table}[htbp]
\caption{Diffusion Index for the Local Attribution Maps displayed in the main paper.}
\label{tab:lam-di}
\centering
\begin{tabular}{lrrrrrr}
\toprule
Method & HAT & SRFormer & MambaIRv2 & PFT & IET & Graphene \\
\midrule
DI & 19.7563 & 12.5254 & 18.6200 & 14.0510 & 15.6248 & 24.9676 \\
\bottomrule
\end{tabular}
\end{table}

\clearpage
\section{Full Benchmark Comparisons}\label{app:benchmarks}

\subsection{Classical SR}\label{app:classical-sr}

\Cref{tab:sr_benchmark_full} provides the full classical SR comparison with SwinIR~\cite{swinir}, CAT-A~\cite{cat}, ART~\cite{art}, HAT~\cite{hat}, SRFormer~\cite{srformer}, IPG~\cite{ipg}, ATD~\cite{atd}, MambaIR~\cite{mambair}, MambaIRv2~\cite{mambairv2}, PFT~\cite{pft}, and IET~\cite{iet}. Graphene uses the datasets, evaluation metrics, and implementation settings described in the main manuscript. With all other settings unchanged, $128\times128$ LR crops raise the PSNR of Graphene by at most $0.08$ dB over $64\times64$ crops, with the largest gain on Urban100. Using Muon rather than Adam alone raises it by at most $0.05$ dB.

\begin{table}[!htbp]
  \caption{Full comparison of Params, FLOPs, inference latency, peak memory, and PSNR/SSIM on \textbf{classical SR} benchmarks. Latency and memory use BF16 compiled inference for all methods ($64\times64$ LR inputs; Section~\ref{sec:setup}); FP32 eager results are in \Cref{tab:fp32_cost}. The best and second-best PSNR/SSIM values at each scale are shown in bold and underlined, respectively; ties receive the same highlighting. Graphene's Params, FLOPs, latency, and memory are bolded to emphasize its efficiency.}
  \label{tab:sr_benchmark_full}
  \centering
  \scriptsize
  \setlength{\tabcolsep}{1.4pt}
  \renewcommand{\arraystretch}{0.94}
  \resizebox{\textwidth}{!}{%
  \begin{tabular}{@{}llrrrrrrrrrrrrrr@{}}
    \toprule
    Scale & Method & Params & FLOPs & Latency & Memory & \multicolumn{2}{c}{Set5} & \multicolumn{2}{c}{Set14} & \multicolumn{2}{c}{B100} & \multicolumn{2}{c}{U100} & \multicolumn{2}{c}{M109} \\
    & & (M) & (G) & (ms) & (MiB) & \multicolumn{1}{c}{PSNR} & \multicolumn{1}{c}{SSIM} & \multicolumn{1}{c}{PSNR} & \multicolumn{1}{c}{SSIM} & \multicolumn{1}{c}{PSNR} & \multicolumn{1}{c}{SSIM} & \multicolumn{1}{c}{PSNR} & \multicolumn{1}{c}{SSIM} & \multicolumn{1}{c}{PSNR} & \multicolumn{1}{c}{SSIM} \\
    \midrule
    $\times 2$ & SwinIR & 11.8 & 102.7 & 2.89 & 85.2 & 38.42 & 0.9623 & 34.46 & 0.9250 & 32.53 & 0.9041 & 33.81 & 0.9427 & 39.92 & 0.9797 \\
    $\times 2$ & CAT-A & 16.5 & 155.0 & 6.77 & 215.3 & 38.51 & 0.9626 & 34.78 & 0.9265 & 32.59 & 0.9047 & 34.26 & 0.9440 & 40.10 & 0.9805 \\
    $\times 2$ & ART & 16.4 & 151.1 & 4.39 & 96.1 & 38.56 & 0.9629 & 34.59 & 0.9267 & 32.58 & 0.9048 & 34.30 & 0.9452 & 40.24 & 0.9808 \\
    $\times 2$ & HAT & 20.6 & 203.4 & 7.10 & 162.2 & 38.63 & 0.9630 & 34.86 & 0.9274 & 32.62 & 0.9053 & 34.45 & 0.9466 & 40.26 & 0.9809 \\
    $\times 2$ & SRFormer & 10.4 & 103.2 & 4.48 & 117.0 & 38.51 & 0.9627 & 34.44 & 0.9253 & 32.57 & 0.9046 & 34.09 & 0.9449 & 40.07 & 0.9802 \\
    $\times 2$ & IPG & 18.1 & 273.6 & 12.68 & 400.6 & 38.61 & 0.9632 & 34.73 & 0.9270 & 32.60 & 0.9052 & 34.48 & 0.9464 & 40.24 & 0.9810 \\
    $\times 2$ & ATD & 21.8 & 258.5 & 8.98 & 121.9 & 38.61 & 0.9629 & 34.92 & 0.9275 & 32.64 & 0.9054 & 34.73 & 0.9476 & 40.35 & 0.9810 \\
    $\times 2$ & MambaIR & 20.4 & 221.0 & 45.18 & 212.9 & 38.57 & 0.9627 & 34.67 & 0.9261 & 32.58 & 0.9048 & 34.15 & 0.9446 & 40.28 & 0.9806 \\
    $\times 2$ & MambaIRv2 & 22.9 & 223.0 & 33.49 & 126.0 & 38.65 & 0.9631 & 34.89 & 0.9275 & 32.62 & 0.9053 & 34.49 & 0.9468 & 40.42 & 0.9810 \\
    $\times 2$ & PFT & 19.6 & 183.5 & 116.23 & 532.1 & 38.68 & 0.9635 & 35.00 & 0.9280 & 32.67 & 0.9058 & 34.90 & 0.9490 & 40.49 & 0.9815 \\
    $\times 2$ & IET & 19.7 & 241.2 & 51.05 & 528.1 & \underline{38.74} & \underline{0.9636} & \underline{35.10} & \underline{0.9286} & \underline{32.71} & \underline{0.9064} & \underline{35.09} & \underline{0.9500} & \underline{40.60} & \underline{0.9816} \\
    \rowcolor{green!5}
    $\times 2$ & \textbf{Graphene} & \textbf{12.9} & \textbf{132.0} & \textbf{8.15} & \textbf{187.9} & \textbf{38.81} & \textbf{0.9637} & \textbf{35.33} & \textbf{0.9300} & \textbf{32.92} & \textbf{0.9082} & \textbf{35.25} & \textbf{0.9512} & \textbf{40.85} & \textbf{0.9820} \\
    \midrule
    $\times 3$ & SwinIR & 11.9 & 104.2 & 2.89 & 85.2 & 34.97 & 0.9318 & 30.93 & 0.8534 & 29.46 & 0.8145 & 29.75 & 0.8826 & 35.12 & 0.9537 \\
    $\times 3$ & CAT-A & 16.6 & 156.5 & 6.77 & 216.3 & 35.06 & 0.9326 & 31.04 & 0.8538 & 29.52 & 0.8160 & 30.12 & 0.8862 & 35.38 & 0.9546 \\
    $\times 3$ & ART & 16.6 & 152.7 & 4.40 & 96.1 & 35.07 & 0.9325 & 31.02 & 0.8541 & 29.51 & 0.8159 & 30.10 & 0.8871 & 35.39 & 0.9548 \\
    $\times 3$ & HAT & 20.8 & 205.0 & 7.12 & 163.5 & 35.07 & 0.9329 & 31.08 & 0.8555 & 29.54 & 0.8167 & 30.23 & 0.8896 & 35.53 & 0.9552 \\
    $\times 3$ & SRFormer & 10.6 & 104.9 & 4.49 & 118.1 & 35.02 & 0.9323 & 30.94 & 0.8540 & 29.48 & 0.8156 & 30.04 & 0.8865 & 35.26 & 0.9543 \\
    $\times 3$ & IPG & 18.3 & 275.2 & 12.67 & 401.2 & 35.10 & 0.9332 & 31.10 & 0.8554 & 29.53 & 0.8168 & 30.36 & 0.8901 & 35.53 & 0.9554 \\
    $\times 3$ & ATD & 22.0 & 260.1 & 8.98 & 123.1 & 35.11 & 0.9330 & 31.13 & 0.8556 & 29.57 & 0.8176 & 30.46 & 0.8917 & 35.63 & 0.9558 \\
    $\times 3$ & MambaIR & 20.6 & 222.6 & 45.19 & 212.9 & 35.08 & 0.9323 & 30.99 & 0.8536 & 29.51 & 0.8157 & 29.93 & 0.8841 & 35.43 & 0.9546 \\
    $\times 3$ & MambaIRv2 & 23.1 & 224.5 & 34.12 & 127.2 & 35.18 & \underline{0.9334} & 31.12 & 0.8557 & 29.55 & 0.8169 & 30.28 & 0.8905 & 35.61 & 0.9556 \\
    $\times 3$ & PFT & 19.8 & 185.0 & 118.32 & 532.5 & 35.15 & 0.9333 & 31.16 & 0.8561 & 29.58 & 0.8178 & 30.56 & 0.8931 & 35.67 & 0.9560 \\
    $\times 3$ & IET & 19.9 & 243.4 & 50.90 & 529.8 & \underline{35.20} & \textbf{0.9337} & \underline{31.23} & \underline{0.8571} & \underline{29.61} & \underline{0.8185} & \underline{30.81} & \underline{0.8966} & \underline{35.82} & \underline{0.9566} \\
    \rowcolor{green!5}
    $\times 3$ & \textbf{Graphene} & \textbf{13.0} & \textbf{133.5} & \textbf{8.22} & \textbf{188.5} & \textbf{35.24} & \textbf{0.9337} & \textbf{31.38} & \textbf{0.8580} & \textbf{29.69} & \textbf{0.8195} & \textbf{31.06} & \textbf{0.8984} & \textbf{35.91} & \textbf{0.9569} \\
    \midrule
    $\times 4$ & SwinIR & 11.9 & 107.7 & 2.93 & 86.3 & 32.92 & 0.9044 & 29.09 & 0.7950 & 27.92 & 0.7489 & 27.45 & 0.8254 & 32.03 & 0.9260 \\
    $\times 4$ & CAT-A & 16.6 & 160.0 & 6.79 & 216.4 & 33.08 & 0.9052 & 29.18 & 0.7960 & 27.99 & 0.7510 & 27.89 & 0.8339 & 32.39 & 0.9285 \\
    $\times 4$ & ART & 16.5 & 156.1 & 4.43 & 97.3 & 33.04 & 0.9051 & 29.16 & 0.7958 & 27.97 & 0.7510 & 27.77 & 0.8321 & 32.31 & 0.9283 \\
    $\times 4$ & HAT & 20.8 & 208.4 & 7.16 & 163.3 & 33.04 & 0.9056 & 29.23 & 0.7973 & 28.00 & 0.7517 & 27.97 & 0.8368 & 32.48 & 0.9292 \\
    $\times 4$ & SRFormer & 10.5 & 108.5 & 4.58 & 118.2 & 32.93 & 0.9041 & 29.08 & 0.7953 & 27.94 & 0.7502 & 27.68 & 0.8311 & 32.21 & 0.9271 \\
    $\times 4$ & IPG & 17.0 & 268.5 & 11.31 & 394.8 & 33.15 & 0.9062 & 29.24 & 0.7973 & 27.99 & 0.7519 & 28.13 & 0.8392 & 32.53 & 0.9300 \\
    $\times 4$ & ATD & 22.0 & 263.5 & 9.00 & 123.0 & 33.14 & 0.9061 & 29.25 & 0.7976 & 28.02 & 0.7524 & 28.22 & 0.8414 & 32.65 & 0.9308 \\
    $\times 4$ & MambaIR & 20.6 & 226.0 & 44.76 & 214.1 & 33.03 & 0.9046 & 29.20 & 0.7961 & 27.98 & 0.7503 & 27.68 & 0.8287 & 32.32 & 0.9272 \\
    $\times 4$ & MambaIRv2 & 23.1 & 228.0 & 33.88 & 126.7 & 33.14 & 0.9057 & 29.23 & 0.7975 & 28.00 & 0.7511 & 27.89 & 0.8344 & 32.57 & 0.9295 \\
    $\times 4$ & PFT & 19.8 & 188.5 & 114.99 & 533.2 & 33.15 & 0.9065 & 29.29 & 0.7978 & 28.02 & 0.7527 & 28.20 & 0.8412 & 32.63 & 0.9306 \\
    $\times 4$ & IET & 19.8 & 248.1 & 51.68 & 530.4 & \underline{33.22} & \underline{0.9069} & \underline{29.35} & \underline{0.7994} & \underline{28.06} & \underline{0.7536} & \underline{28.43} & \underline{0.8464} & \underline{32.81} & \textbf{0.9318} \\
    \rowcolor{green!5}
    $\times 4$ & \textbf{Graphene} & \textbf{13.0} & \textbf{137.0} & \textbf{8.34} & \textbf{192.6} & \textbf{33.27} & \textbf{0.9070} & \textbf{29.43} & \textbf{0.8001} & \textbf{28.12} & \textbf{0.7542} & \textbf{28.67} & \textbf{0.8483} & \textbf{32.89} & \underline{0.9316} \\
    \bottomrule
  \end{tabular}%
  }
\end{table}

\clearpage
\subsection{Lightweight SR}\label{app:lightweight-sr}

The lightweight SR experiments use the datasets, evaluation metrics, and implementation settings described in the main manuscript. Graphene-light is trained on DF2K, as is Graphene; against training on DIV2K, DF2K raises its PSNR by at most $0.23$ dB. With DIV2K training, as used by IET-light, Graphene-light therefore remains at least $0.15$ dB above IET-light on Urban100 at $\times4$. \Cref{tab:lightweight_sr} provides the full lightweight SR comparison. Graphene-light uses fewer than one million parameters at each scale and achieves the highest PSNR in 12 of the 15 dataset--scale comparisons. On Urban100 at $\times4$, it outperforms IET-light~\cite{iet} by \textbf{0.38 dB} while using \textbf{13.7\% fewer parameters}, \textbf{44.4\% fewer FLOPs}, \textbf{\lightLatencyReduction\% lower latency}, and \textbf{\lightMemoryReduction\% lower peak memory}, demonstrating Graphene’s scalability to a sub-million-parameter configuration.

The comparison includes SwinIR-light~\cite{swinir}, SRFormer-light~\cite{srformer}, ATD-light~\cite{atd}, MambaIR-light~\cite{mambair}, MambaIRv2-light~\cite{mambairv2}, PFT-light~\cite{pft}, UCAN~\cite{ucan}, and IET-light~\cite{iet}.

\begin{table}[!htbp]
  \caption{Full comparison of Params, FLOPs, inference latency, peak memory, and PSNR/SSIM on \textbf{lightweight SR} benchmarks. Latency and memory use BF16 compiled inference for all methods ($64\times64$ LR inputs; Section~\ref{sec:setup}); FP32 eager results are in \Cref{tab:fp32_cost}. The best and second-best PSNR/SSIM values at each scale are shown in bold and underlined, respectively; ties receive the same highlighting. Graphene-light's Params, FLOPs, latency, and memory are bolded to emphasize its efficiency.}
  \label{tab:lightweight_sr}
  \centering
  \scriptsize
  \setlength{\tabcolsep}{1.4pt}
  \renewcommand{\arraystretch}{0.94}
  \resizebox{\textwidth}{!}{%
  \begin{tabular}{@{}llrrrrrrrrrrrrrr@{}}
    \toprule
    Scale & Method & Params & FLOPs & Latency & Memory & \multicolumn{2}{c}{Set5} & \multicolumn{2}{c}{Set14} & \multicolumn{2}{c}{B100} & \multicolumn{2}{c}{U100} & \multicolumn{2}{c}{M109} \\
    & & (K) & (G) & (ms) & (MiB) & \multicolumn{1}{c}{PSNR} & \multicolumn{1}{c}{SSIM} & \multicolumn{1}{c}{PSNR} & \multicolumn{1}{c}{SSIM} & \multicolumn{1}{c}{PSNR} & \multicolumn{1}{c}{SSIM} & \multicolumn{1}{c}{PSNR} & \multicolumn{1}{c}{SSIM} & \multicolumn{1}{c}{PSNR} & \multicolumn{1}{c}{SSIM} \\
    \midrule
    $\times 2$ & SwinIR-light & 910.0 & 8.7 & 1.13 & 26.1 & 38.14 & 0.9611 & 33.86 & 0.9206 & 32.31 & 0.9012 & 32.76 & 0.9340 & 39.12 & 0.9783 \\
    $\times 2$ & SRFormer-light & 853.0 & 8.2 & 1.55 & 28.2 & 38.23 & 0.9613 & 33.94 & 0.9209 & 32.36 & 0.9019 & 32.91 & 0.9353 & 39.28 & 0.9785 \\
    $\times 2$ & ATD-light & 761.0 & 13.6 & 3.73 & 19.1 & 38.29 & 0.9616 & 34.10 & 0.9217 & 32.39 & 0.9023 & 33.27 & 0.9375 & 39.52 & 0.9789 \\
    $\times 2$ & MambaIR-light & 905.0 & 11.9 & 28.96 & 24.8 & 38.16 & 0.9610 & 34.00 & 0.9212 & 32.34 & 0.9017 & 32.92 & 0.9356 & 39.31 & 0.9779 \\
    $\times 2$ & MambaIRv2-light & 774.0 & 10.0 & 18.23 & 19.3 & 38.26 & 0.9615 & 34.09 & 0.9221 & 32.36 & 0.9019 & 33.26 & 0.9378 & 39.35 & 0.9785 \\
    $\times 2$ & PFT-light & 776.0 & 10.2 & 44.64 & 294.2 & \underline{38.36} & \underline{0.9620} & 34.19 & 0.9232 & 32.43 & 0.9030 & 33.67 & 0.9411 & 39.55 & 0.9792 \\
    $\times 2$ & UCAN & 689.0 & 11.2 & 19.58 & 145.4 & 38.34 & 0.9618 & 34.27 & 0.9242 & 32.39 & 0.9025 & 33.22 & 0.9379 & 39.54 & 0.9790 \\
    $\times 2$ & IET-light & 783.0 & 12.4 & 37.38 & 256.5 & \textbf{38.44} & \textbf{0.9624} & \underline{34.28} & \underline{0.9245} & \underline{32.50} & \underline{0.9038} & \textbf{34.03} & \textbf{0.9435} & \underline{39.75} & \underline{0.9794} \\
    \rowcolor{green!5}
    $\times 2$ & \textbf{Graphene-light} & \textbf{672.1} & \textbf{6.8} & \textbf{1.66} & \textbf{20.2} & 38.34 & 0.9617 & \textbf{34.59} & \textbf{0.9253} & \textbf{32.58} & \textbf{0.9042} & \underline{33.91} & \underline{0.9424} & \textbf{39.90} & \textbf{0.9796} \\
    \midrule
    $\times 3$ & SwinIR-light & 918.0 & 8.8 & 1.12 & 26.4 & 34.62 & 0.9289 & 30.54 & 0.8463 & 29.20 & 0.8082 & 28.66 & 0.8624 & 33.98 & 0.9478 \\
    $\times 3$ & SRFormer-light & 861.0 & 8.3 & 1.53 & 28.4 & 34.67 & 0.9296 & 30.57 & 0.8469 & 29.26 & 0.8099 & 28.81 & 0.8655 & 34.19 & 0.9489 \\
    $\times 3$ & ATD-light & 767.0 & 13.7 & 3.79 & 19.4 & 34.74 & 0.9300 & 30.68 & 0.8485 & 29.32 & 0.8109 & 29.17 & 0.8709 & 34.60 & 0.9506 \\
    $\times 3$ & MambaIR-light & 913.0 & 11.9 & 28.61 & 25.1 & 34.72 & 0.9296 & 30.63 & 0.8475 & 29.29 & 0.8099 & 29.00 & 0.8689 & 34.39 & 0.9495 \\
    $\times 3$ & MambaIRv2-light & 781.0 & 10.0 & 18.17 & 19.6 & 34.71 & 0.9298 & 30.68 & 0.8483 & 29.26 & 0.8098 & 29.01 & 0.8689 & 34.41 & 0.9497 \\
    $\times 3$ & PFT-light & 783.0 & 10.2 & 46.46 & 294.4 & 34.81 & 0.9305 & 30.75 & 0.8493 & 29.33 & 0.8116 & 29.43 & 0.8759 & 34.60 & 0.9510 \\
    $\times 3$ & UCAN & 696.0 & 11.3 & 20.33 & 145.7 & 34.83 & \underline{0.9308} & 30.72 & 0.8493 & 29.32 & 0.8121 & 29.15 & 0.8712 & 34.62 & 0.9508 \\
    $\times 3$ & IET-light & 790.0 & 12.5 & 38.19 & 256.9 & \textbf{34.90} & \textbf{0.9314} & \underline{30.83} & \underline{0.8504} & \underline{29.41} & \underline{0.8135} & \underline{29.75} & \underline{0.8808} & \underline{34.89} & \textbf{0.9524} \\
    \rowcolor{green!5}
    $\times 3$ & \textbf{Graphene-light} & \textbf{680.3} & \textbf{6.9} & \textbf{1.59} & \textbf{20.4} & \underline{34.87} & 0.9305 & \textbf{31.00} & \textbf{0.8511} & \textbf{29.49} & \textbf{0.8140} & \textbf{29.98} & \textbf{0.8809} & \textbf{35.01} & \underline{0.9520} \\
    \midrule
    $\times 4$ & SwinIR-light & 930.0 & 8.8 & 1.12 & 26.8 & 32.44 & 0.8976 & 28.77 & 0.7858 & 27.69 & 0.7406 & 26.47 & 0.7980 & 30.92 & 0.9151 \\
    $\times 4$ & SRFormer-light & 873.0 & 8.4 & 1.53 & 28.8 & 32.51 & 0.8988 & 28.82 & 0.7872 & 27.73 & 0.7422 & 26.67 & 0.8032 & 31.17 & 0.9165 \\
    $\times 4$ & ATD-light & 777.0 & 13.7 & 3.80 & 19.8 & 32.63 & 0.8998 & 28.89 & 0.7886 & 27.79 & 0.7440 & 26.97 & 0.8107 & 31.48 & 0.9198 \\
    $\times 4$ & MambaIR-light & 924.0 & 12.0 & 28.32 & 25.5 & 32.51 & 0.8993 & 28.85 & 0.7876 & 27.75 & 0.7423 & 26.75 & 0.8051 & 31.26 & 0.9175 \\
    $\times 4$ & MambaIRv2-light & 790.0 & 10.1 & 18.46 & 19.9 & 32.51 & 0.8992 & 28.84 & 0.7878 & 27.75 & 0.7426 & 26.82 & 0.8079 & 31.24 & 0.9182 \\
    $\times 4$ & PFT-light & 792.0 & 10.3 & 45.12 & 294.8 & 32.63 & 0.9005 & 28.92 & 0.7891 & 27.79 & 0.7445 & 27.20 & 0.8171 & 31.51 & 0.9204 \\
    $\times 4$ & UCAN & 705.0 & 11.3 & 20.47 & 146.0 & 32.65 & \underline{0.9010} & 28.95 & 0.7899 & 27.79 & 0.7454 & 26.89 & 0.8097 & 31.50 & 0.9200 \\
    $\times 4$ & IET-light & 801.0 & 12.6 & 38.52 & 258.8 & \underline{32.70} & \textbf{0.9019} & \underline{29.00} & \underline{0.7908} & \underline{27.85} & \underline{0.7464} & \underline{27.41} & \underline{0.8227} & \underline{31.73} & \textbf{0.9226} \\
    \rowcolor{green!5}
    $\times 4$ & \textbf{Graphene-light} & \textbf{691.6} & \textbf{7.0} & \textbf{1.60} & \textbf{20.8} & \textbf{32.80} & 0.9007 & \textbf{29.19} & \textbf{0.7922} & \textbf{27.94} & \textbf{0.7468} & \textbf{27.79} & \textbf{0.8243} & \textbf{31.93} & \underline{0.9225} \\
    \bottomrule
  \end{tabular}%
  }
\end{table}

\subsection{Parameter Count, FLOPs, Latency, and Peak Memory}\label{app:efficiency}

Efficiency measurements use an RGB tensor of shape $1\times3\times64\times64$. Parameter counts include the reconstruction head for the corresponding scale. For Graphene and Graphene-light profiling, deterministic attention and IRB pair-mask buffers are constructed for the $64\times64$ input without changing learned weights. The MAC-based operation counter uses two FLOPs per multiply--accumulate and model-specific handlers for supported operations.

FP32 eager inference uses the GPU, software stack, benchmark script, and timing protocol of Section~\ref{sec:setup}, with TF32 disabled and cuDNN autotuning enabled. \Cref{tab:fp32_cost} reports FP32 eager latency and peak memory for every method at $\times2$, $\times3$, and $\times4$. On Urban100 at $\times4$, Graphene exceeds IET by $0.24$ dB, using $34.3\%$ fewer parameters, $44.8\%$ fewer FLOPs, $19.2\%$ lower latency, and $60.6\%$ lower peak memory; Graphene-light exceeds IET-light by $0.38$ dB, using $13.7\%$ fewer parameters, $44.4\%$ fewer FLOPs, $22.4\%$ lower latency, and $79.9\%$ lower peak memory. IET and IET-light use NATTEN, the neighborhood-attention library required by their released code, restricted to its FP32 kernels. Fresh processes settle on latency levels that differ by about 5\%; latency and peak memory are therefore the medians over seven runs at each scale. In both settings, relative reductions are computed before rounding. The benchmark process runs on the GPU's NUMA-local CPU cores; runs that overlap another process on the same GPU or measurable activity from other processes on those cores are discarded and repeated. Peak memory includes model parameters, inputs, buffers, and peak allocated activations; it is distinct from reserved allocator memory or process memory reported by a system monitor. PFT retains the cache-clearing calls in its released forward path. The released benchmark package reproduces these measurements.

\begin{figure}[!htbp]
\centering
\includegraphics[width=\linewidth]{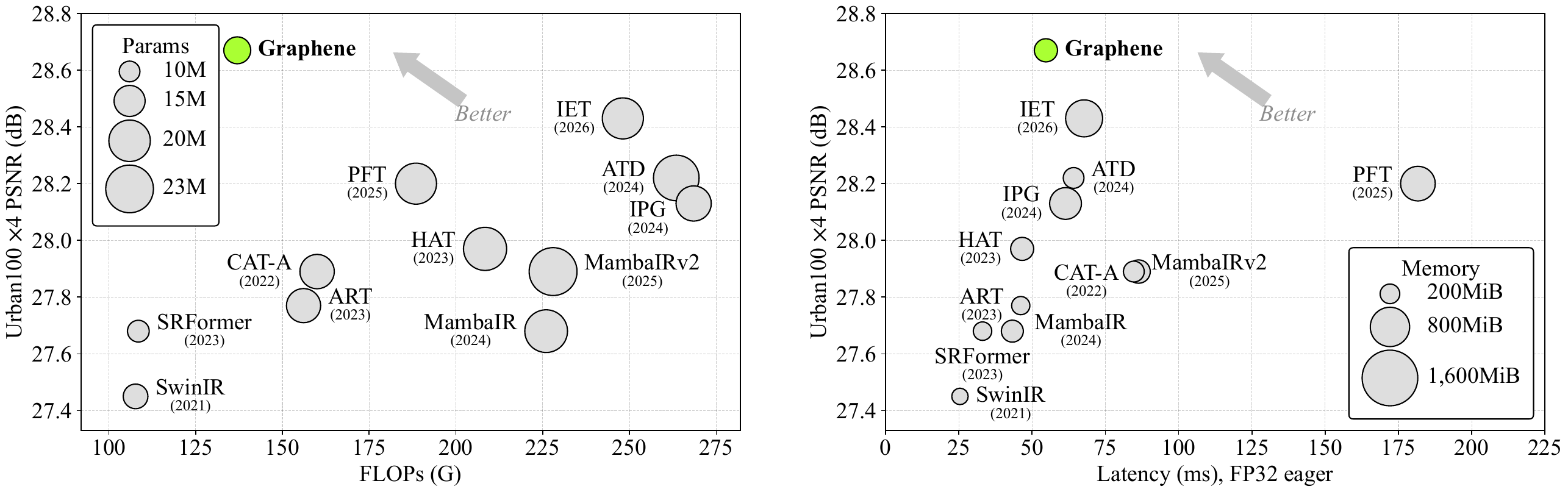}
\vspace{-2.5em}
\caption{FP32 eager counterpart of \Cref{fig:bubble}. \textit{Left:} PSNR versus FLOPs, with circle diameters indicating parameter counts. \textit{Right:} PSNR versus FP32 eager latency with TF32 disabled for every method, with circle areas indicating peak memory; IET uses NATTEN's FP32 kernels. Values match \Cref{tab:fp32_cost}.}
\label{fig:fp32_eager}
\end{figure}

\makeatletter\setlength{\@fptop}{0pt}\makeatother
\begin{table}[!htb]
\caption{FP32 eager latency and peak memory at $64\times64$ LR inputs, with TF32 disabled; IET and IET-light use NATTEN's FP32 kernels. Each value is the median over seven runs.}
\label{tab:fp32_cost}
\centering
\small
\renewcommand{\arraystretch}{0.9}
\begin{tabular}{lrrrrrr}
\toprule
& \multicolumn{2}{c}{$\times2$} & \multicolumn{2}{c}{$\times3$} & \multicolumn{2}{c}{$\times4$} \\
\cmidrule(lr){2-3}\cmidrule(lr){4-5}\cmidrule(lr){6-7}
Method & Latency & Memory & Latency & Memory & Latency & Memory \\
& (ms) & (MiB) & (ms) & (MiB) & (ms) & (MiB) \\
\midrule
\multicolumn{7}{c}{\textbf{Classical SR}} \\
\midrule
SwinIR~\cite{swinir} & 25.6 & 138.5 & 25.8 & 138.9 & 25.4 & 139.6 \\
CAT-A~\cite{cat} & 92.5 & 281.5 & 82.3 & 282.8 & 86.3 & 282.1 \\
ART~\cite{art} & 42.9 & 170.9 & 43.4 & 170.9 & 46.1 & 172.1 \\
HAT~\cite{hat} & 49.1 & 273.1 & 48.2 & 274.3 & 46.6 & 274.2 \\
SRFormer~\cite{srformer} & 32.5 & 175.1 & 32.1 & 175.9 & 33.1 & 176.3 \\
IPG~\cite{ipg} & 74.1 & 529.7 & 72.3 & 530.6 & 61.4 & 522.9 \\
ATD~\cite{atd} & 64.2 & 222.1 & 64.5 & 223.0 & 64.2 & 222.8 \\
MambaIR~\cite{mambair} & 44.9 & 248.7 & 43.5 & 248.6 & 43.3 & 249.8 \\
MambaIRv2~\cite{mambairv2} & 86.5 & 223.4 & 86.2 & 224.1 & 84.7 & 224.6 \\
PFT~\cite{pft} & 180.4 & 621.2 & 177.8 & 621.9 & 181.6 & 621.8 \\
IET~\cite{iet} & 67.4 & 703.6 & 67.6 & 705.3 & 67.7 & 705.9 \\
\rowcolor{green!5}
\textbf{Graphene} & \textbf{53.7} & \textbf{276.9} & \textbf{54.9} & \textbf{276.9} & \textbf{54.7} & \textbf{278.1} \\
\midrule
\multicolumn{7}{c}{\textbf{Lightweight SR}} \\
\midrule
SwinIR~\cite{swinir} & 16.8 & 71.0 & 16.3 & 71.3 & 18.1 & 71.7 \\
SRFormer~\cite{srformer} & 22.6 & 71.0 & 20.8 & 71.3 & 23.4 & 71.7 \\
ATD~\cite{atd} & 49.4 & 79.6 & 49.7 & 79.9 & 48.7 & 80.3 \\
MambaIR~\cite{mambair} & 28.4 & 67.3 & 29.1 & 67.6 & 28.8 & 68.0 \\
MambaIRv2~\cite{mambairv2} & 48.9 & 86.7 & 49.5 & 86.9 & 50.7 & 87.3 \\
PFT~\cite{pft} & 110.1 & 325.9 & 109.7 & 326.1 & 112.6 & 326.5 \\
UCAN~\cite{ucan} & 73.2 & 194.8 & 74.0 & 195.0 & 74.3 & 195.4 \\
IET~\cite{iet} & 28.4 & 345.6 & 30.8 & 346.0 & 31.1 & 347.4 \\
\rowcolor{green!5}
\textbf{Graphene-light} & \textbf{25.3} & \textbf{69.3} & \textbf{24.0} & \textbf{69.6} & \textbf{24.1} & \textbf{69.9} \\
\bottomrule
\end{tabular}
\end{table}

\begin{table}[!t]
\caption{$\times4$ FP32 eager latency and peak memory at larger LR inputs (width$\times$height), with TF32 disabled; IET and IET-light use NATTEN's FP32 kernels. Each value is the median over three runs of 5 warm-up and 20 timed passes; OOM: out of GPU memory.}
\label{tab:fp32_large}
\centering
\small
\renewcommand{\arraystretch}{0.9}
\begin{tabular}{lrrrr}
\toprule
& \multicolumn{2}{c}{$320\times180$} & \multicolumn{2}{c}{$1280\times720$} \\
\cmidrule(lr){2-3}\cmidrule(lr){4-5}
Method & Latency (ms) & Memory (GiB) & Latency (s) & Memory (GiB) \\
\midrule
\multicolumn{5}{c}{\textbf{Classical SR}} \\
\midrule
SwinIR~\cite{swinir} & 108.2 & 0.69 & 1.67 & 9.31 \\
CAT-A~\cite{cat} & 300.8 & 2.86 & OOM & OOM \\
ART~\cite{art} & 377.6 & 10.72 & OOM & OOM \\
HAT~\cite{hat} & 187.4 & 2.41 & 3.24 & 34.48 \\
SRFormer~\cite{srformer} & 170.0 & 0.85 & 2.59 & 10.59 \\
IPG~\cite{ipg} & 485.7 & 5.18 & 8.41 & 76.87 \\
ATD~\cite{atd} & 260.4 & 1.66 & 4.03 & 23.23 \\
MambaIR~\cite{mambair} & 289.2 & 2.10 & 4.84 & 29.94 \\
MambaIRv2~\cite{mambairv2} & 366.1 & 1.57 & 6.79 & 21.88 \\
PFT~\cite{pft} & 1433.7 & 7.17 & 22.47 & 107.87 \\
IET~\cite{iet} & 686.9 & 6.18 & 11.71 & 92.85 \\
\rowcolor{green!5}
\textbf{Graphene} & \textbf{229.5} & \textbf{1.88} & \textbf{4.03} & \textbf{26.54} \\
\midrule
\multicolumn{5}{c}{\textbf{Lightweight SR}} \\
\midrule
SwinIR~\cite{swinir} & 65.5 & 0.38 & 0.53 & 5.14 \\
SRFormer~\cite{srformer} & 75.5 & 0.36 & 0.58 & 4.63 \\
ATD~\cite{atd} & 58.5 & 0.67 & 0.67 & 9.46 \\
MambaIR~\cite{mambair} & 40.4 & 0.49 & 0.63 & 6.76 \\
MambaIRv2~\cite{mambairv2} & 68.4 & 0.77 & 0.87 & 10.92 \\
PFT~\cite{pft} & 617.1 & 4.13 & 8.27 & 62.59 \\
UCAN~\cite{ucan} & 78.6 & 1.77 & 0.79 & 25.79 \\
IET~\cite{iet} & 184.9 & 3.09 & 2.91 & 46.45 \\
\rowcolor{green!5}
\textbf{Graphene-light} & \textbf{27.4} & \textbf{0.50} & \textbf{0.44} & \textbf{6.88} \\
\bottomrule
\end{tabular}
\end{table}

The FLOPs, latency, and peak memory in \Cref{tab:sr_benchmark,tab:sr_benchmark_full,tab:lightweight_sr,tab:fp32_cost} apply to $64\times64$ LR inputs; operation counts and intermediate storage scale linearly with image area at fixed window size. \Cref{tab:fp32_large} reports $\times4$ FP32 eager results at $320\times180$ and $1280\times720$ LR inputs for every method. Graphene and Graphene-light are $2.9$--$6.8\times$ faster than IET and IET-light with $3.3$--$6.7\times$ less memory; faster or leaner models are $\geq0.45$ dB lower on Urban100. Graphene-light is the fastest lightweight model at both sizes, with peak memory above SwinIR-light, SRFormer-light, and MambaIR-light and below the others. SwinIR, SRFormer, and HAT are faster than Graphene at both sizes, ATD is as fast at $1280\times720$, and SwinIR, SRFormer, MambaIRv2, and ATD use less memory. CAT-A and ART exceed GPU memory at $1280\times720$. \Cref{tab:large_input} uses the BF16 compiled settings and code changes of Section~\ref{sec:setup} and reports the fastest variant at each size. Matrix products with $2^{31}$ or more elements in an operand or the output use cuBLAS, because the Triton matrix-multiplication templates of Inductor index with 32-bit offsets. At $1280\times720$, ART exceeds GPU memory in every compilation mode, CAT-A runs without CUDA graphs because max-autotune exceeds GPU memory, and IET uses NATTEN's FP32 kernels because its GEMM kernels fail at this size.

\end{document}